\documentclass[journal]{IEEEtran}
\usepackage{ifpdf}
\usepackage{cite}
\usepackage{xcolor}
\ifCLASSINFOpdf
  \usepackage[pdftex]{graphicx}
\else
  \usepackage[dvips]{graphicx}
\fi
\usepackage{mathtools}
\usepackage{extarrows}
\usepackage{algorithmic}
\usepackage{array}
\ifCLASSOPTIONcompsoc
  \usepackage[caption=false,font=normalsize,labelfont=sf,textfont=sf]{subfig}
\else
  \usepackage[caption=false,font=footnotesize]{subfig}
\fi
\usepackage{url}
\usepackage{tabularx}
\usepackage{multirow}
\usepackage{amssymb}
\usepackage{threeparttable,booktabs}
\usepackage{hyperref}
\usepackage{floatrow}
\usepackage{diagbox}
\newfloatcommand{capbtabbox}{table}[][\FBwidth]
\usepackage{blindtext}

\definecolor{LightGrey}{rgb}{0.9,0.9,0.9}
\definecolor{my_orange}{RGB}{255,153,51}
\definecolor{my_red}{RGB}{234,107,102}
\definecolor{applegreen}{rgb}{0.0, 0.5, 0.0}
\definecolor{edscwblue}{HTML}{2988bc}
\definecolor{emred}{HTML}{ae4951}
\definecolor{bblue}{HTML}{0063c8}
\newcommand\tstrut{\rule{0pt}{2.4ex}}
\newcommand\bstrut{\rule[-1.0ex]{0pt}{0pt}}

\begin{document}
\title{AdaPool: Exponential Adaptive Pooling for Information-Retaining Downsampling}
%
%
%

\author{Alexandros~Stergiou, \textit{Student Member}, IEEE,
        Ronald~Poppe, \textit{Senior Member}, IEEE
\thanks{
This work was done while A. Stergiou was with Utrecht University.
\newline
A. Stergiou is with the Department of Computer Science, University of Bristol, Bristol, BS8 1UB, United Kingdom, e-mail: alexandros.stergiou@bristol.ac.uk 
\newline
R. Poppe is with the Department
of Information and Computing Sciences, Utrecht University, Utrecht, 3584 CC, The Netherlands, e-mail: r.w.poppe@uu.nl \newline
\newline
Code URL: \url{https://git.io/JcDHN} \newline
Dataset URL: \url{https://alexandrosstergiou.github.io/datasets/Inter4K/}
}}

\markboth{}%
{}


\maketitle
\begin{abstract}
Pooling layers are essential building blocks of convolutional neural networks (CNNs), to reduce computational overhead and increase the receptive fields of proceeding convolutional operations. Their goal is to produce downsampled volumes that closely resemble the input volume while, ideally, also being computationally and memory efficient. Meeting both these requirements remains a challenge. To this end, we propose an adaptive and exponentially weighted pooling method: \textit{adaPool}. Our method learns a regional-specific fusion of two sets of pooling kernels that are based on the exponent of the Dice-S\o rensen coefficient and the exponential maximum, respectively. AdaPool improves the preservation of detail on a range of tasks including image and video classification and object detection. A key property of adaPool is its bidirectional nature. In contrast to common pooling methods, the learned weights can also be used to upsample activation maps. We term this method adaUnPool. We evaluate \textit{adaUnPool} on image and video super-resolution and frame interpolation. For benchmarking, we introduce \textit{Inter4K}, a novel high-quality, high frame-rate video dataset. Our experiments demonstrate that adaPool systematically achieves better results across tasks and backbones, while introducing a minor additional computational and memory overhead.
\end{abstract}

\begin{IEEEkeywords}
pooling, downsampling, upsampling
\end{IEEEkeywords}

\IEEEpeerreviewmaketitle

\section{Introduction}
\label{sec:intro}

Pooling methods downsample spatial input to a lower resolution. Their goal is to minimize the computational overhead of subsequent network operations and to increase their receptive fields. Pooling operations are essential in image and video processing approaches, including those based on CNNs. An important aspect of pooling is that it introduces a loss of information within the model. Thus, the retainment of detail in the structural aspects of the input, such as contrast and texture, can become challenging. As pooling is a key component in virtually all popular CNN architectures, it is necessary to ensure that this information loss does not incur a cost in performance.

\begin{figure}[t]
\includegraphics[width=\linewidth]{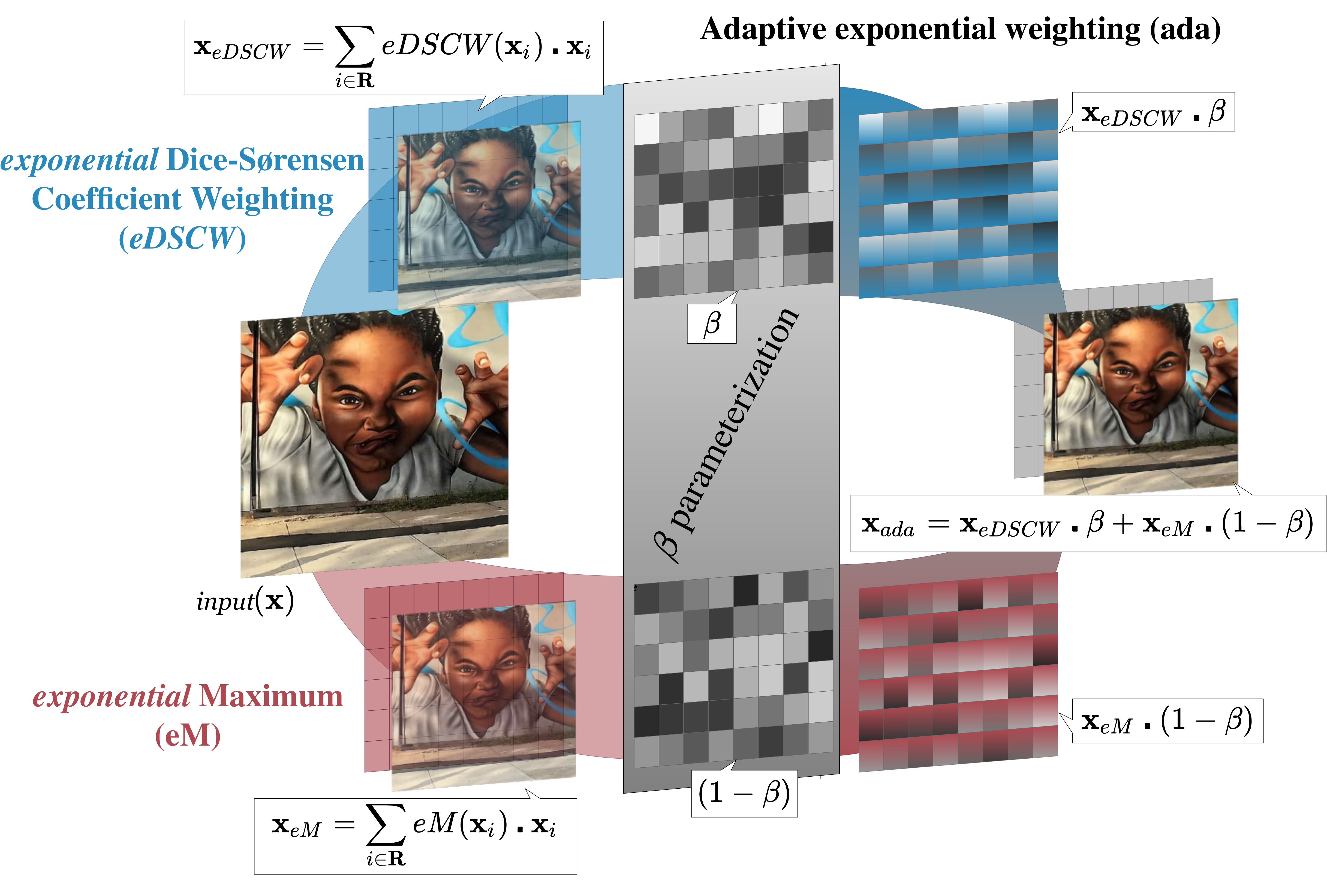}
\caption{\textbf{AdaPool downsampling}. The output is the combination of two processes. The first uses exponential Dice-S\o rensen Coefficient Weighting (\textcolor{edscwblue}{\textit{e}DSCW}) downsampling, based on a region's mean ($\mathbf{\overline x}$). The second downsamples using the exponential maximum (\textcolor{emred}{\textit{e}M}). Both outputs  (\textbf{x}$_{\textit{e}M}$,\textbf{x}$_{\textit{e}DSCW}$) are summed with region-based weight masks $\boldsymbol{\beta}$ and (1-$\boldsymbol{\beta}$) to produce the adaptively weighted output (\textbf{x}$_{ada}$).}
\label{fig:adapool_formulation}
\end{figure}

A range of pooling methods has been proposed, each with different properties (see Section~\ref{sec:related}). Most architectures use maximum or average pooling, both of which are fast and memory efficient but leave room for improvement in terms of retaining information. Other approaches use trainable sub-networks. Such methods have shown some improvements over average or maximum pooling, but they are typically less efficient and not generally applicable because their parameters need to be determined beforehand.


\begin{figure*}[ht]
\includegraphics[trim={0 6cm 0 0},width=\linewidth]{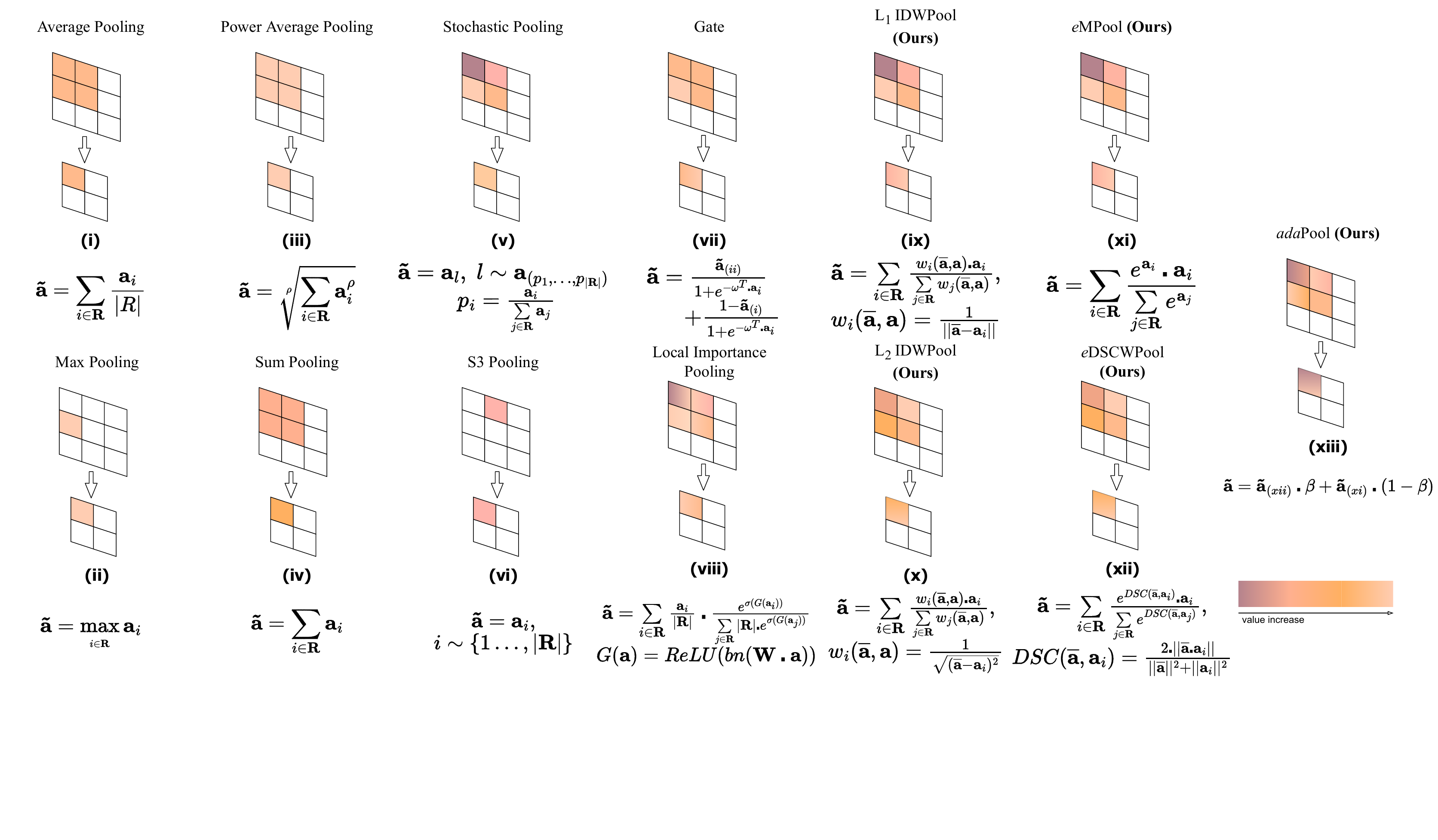}
\caption{\textbf{Pooling variants}. $\mathbf{R}$ denotes the kernel neighborhood as a set of pixels. (i-ii) \textbf{Average} and \textbf{maximum pooling} are based on the average or maximum activation value of the kernel region. (iii) \textbf{Power-average pooling}~\cite{estrach2014signal,gulcehre2014learned} is proportional to average pooling raised to the power of $\rho$. When $\rho \! \rightarrow \!\infty$ the output equals maximum pooling, while $\rho \! = \! 1$ equals average pooling. (iv) \textbf{Sum pooling} is also proportional to average pooling with all kernel activations summed in the output. (v) \textbf{Stochastic pooling}~\cite{zeiler2013stochastic} samples a random activation from the kernel region. (vi) \textbf{Stochastic Spatial Sampling} (S3Pool)~\cite{zhai2017s3pool} samples horizontal and vertical regions given a specified stride. (vi) \textbf{Gate pooling}~\cite{lee2016generalizing} uses max-average pooling based on a gating mask ($\omega$) and a sigmoid function. (viii) \textbf{Local Importance Pooling} (LIP)~\cite{gao2019lip} uses a trainable sub-net $G$ to enhance specific features. (ix-x) \textbf{L$_{1}$ and L$_{2}$ Inverse Distance Weighting Pooling} (IDW, ours) weighs kernel regions based on their inverse distance to the mean activation ($\mathbf{\overline a}$). (xi) \textbf{Exponential maximum pooling} (\textit{e}mPool/SoftPool, ours) exponentially weighs activations using a softmax kernel. (xii) \textbf{Exponential Dice-S\o rensen Coefficient Weighting Pooling} (\textit{e}DSCWPool, ours) uses the exponent Dice-S\o rensen Coefficient \cite{dice1945measures,sorensen1948method} of the kernel activations ($\mathbf{a}_{i}$) and their average ($\mathbf{\overline a}$) as weights. (xiii) \textbf{Adaptive exponential pooling} (adaPool, ours) combines (xi) and (xii) with a trainable mask of weights $\boldsymbol{\beta}$.}
\label{fig:pooling_variants}
\end{figure*}

In this work, we study how the shortcomings of pooling methods can be addressed with low-computational approaches based on exponential weighting. We introduce methods to weigh kernel regions, either based on the softmax-weighted sum of activations \cite{stergiou2021refining}, or based on the exponent of the similarity between each activation and the mean activation within the kernel region obtained by the Dice-S\o rensen Coefficient \cite{dice1945measures,sorensen1948method}. We then propose \textit{adaPool} as the learned fusion of both methods, schematically visualized in Figure~\ref{fig:adapool_formulation}. AdaPool does not average over high-frequency patterns as in average pooling, nor does it focus exclusively on such patterns as in maximum pooling. Instead, adaPool provides a balance between retaining informative detail and the local image structure.

Many tasks, including instance segmentation, image generation and super-resolution, require upsampling of inputs or signals, which has the inverse goal of pooling. With the exception of LiftPool~\cite{zhao2021liftpool}, pooling operations cannot be reversed as this would lead to sparse upsampling results (e.g., using maximum pooling~\cite{badrinarayanan2017segnet}). Common upsampling approaches such as interpolation, transposed convolutions and de-convolution approximate, rather than reconstruct, the higher-resolution features. The lack of inclusion of prior knowledge is an obstacle as the encoding of information to a lower resolution comes at a loss of local information. Instead, we argue that introducing prior local knowledge benefits the upsampling process. Based on the same formulation as adaPool, we propose \textit{adaUnPool} for upsampling.

We demonstrate the favorable effects of adaPool in preserving descriptive features. Consequently, this allows models with adaPool to consistently improve classification and recognition performance. AdaPool maintains a low computational cost and provides an approach to retain prior information. We further introduce adaUnPool and address super-resolution and interpolation tasks. Summarized, we make the following contributions:
\begin{itemize}
    \item We adapt Inverse Distance Weighting (IDW) \cite{shepard1968two} for pooling and extend it by using a similarity measure through the Dice-S\o rensen Coefficient (DSC), by utilizing its exponent \textit{e}DSC to weigh kernel elements.
    \item We propose adaPool, a parameterized learnable fusion of portions from the smooth approximation of the maximum and average. Using the inverse formulation, we develop upsampling process adaUnPool.
    \item We introduce a collection of 1,000 4K videos with high frame-rates, \textit{Inter4K}, to benchmark frame super-resolution and interpolation algorithms.
    \item We experiment on multiple global and local-based tasks including image and video classification, and object detection. We show consistent improvements by replacing original pooling layers with adaPool. We also demonstrate the improved performance of adaUnPool on image and video super-resolution and video frame interpolation.
\end{itemize}

The remainder of the paper is structured as follows. We first discuss related work. We then detail our downsampling methods \textit{e}DSCPool, \textit{e}MPool, and adaPool as well as upsampling method adaUnPool (Section~\ref{sec:method}). We introduce Inter4K in Section~\ref{sec:inter4k} and evaluate on global and local-based image and video tasks (Section~\ref{sec:experiments}). We conclude in Section~\ref{sec:conclusion}.

\begin{figure*}[t]
\includegraphics[width=\linewidth]{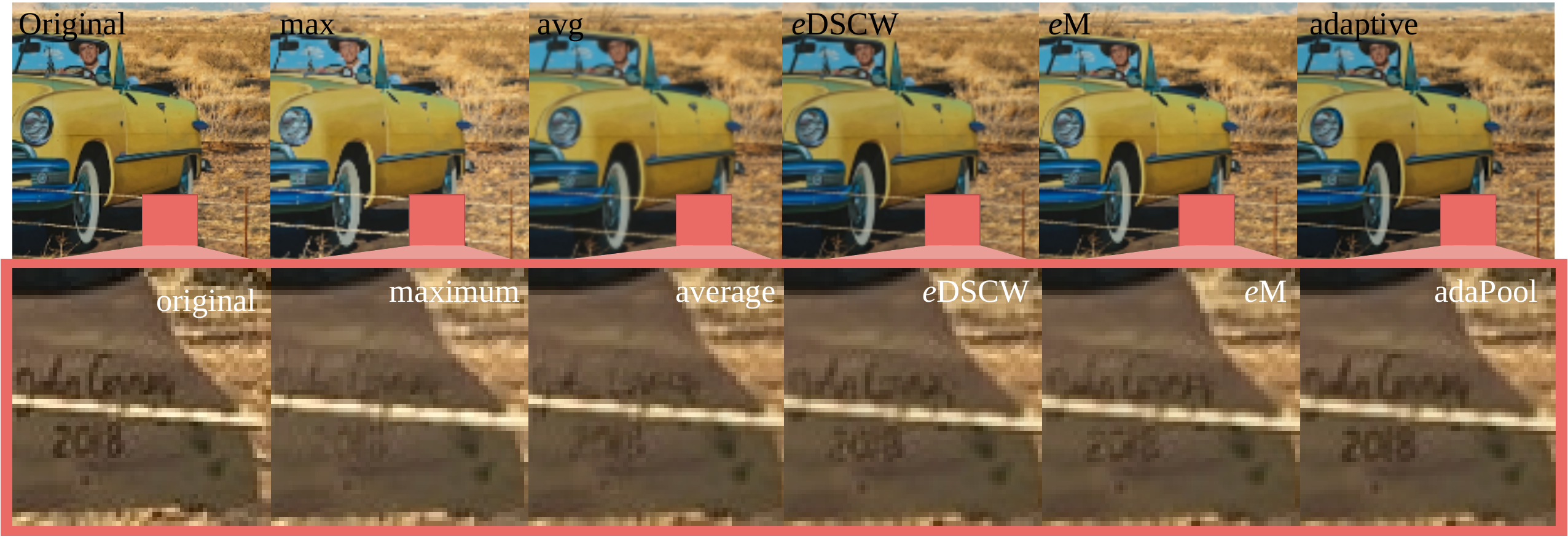}
\caption{\textbf{Example of detail preservation with different pooling methods}. Common methods such as average and maximum pooling result in a distorted signature with unrecognizable details such as numbers or characters. Exponential weighting through either normalized local maximum (\textit{e}M) or similarity-based measures (\textit{e}DSCW) better capture details. Further improvements in the detail and representation quality are observed when introducing an adaptive fusion between both of these exponential weighting methods (adaPool).}
\label{fig:adapool_detail}
\end{figure*}

\section{Related Work}
\label{sec:related}

\textbf{Pooling hand-crafted features}. Downsampling has been widely used in hand-coded feature extraction. In Bag-of-Words (BoW,~\cite{csurka2004visual}), images are represented as groups of local patches that are pooled and then encoded as vectors \cite{wang2010locality}. Based on this approach, Spatial Pyramid Matching (SPM) \cite{lazebnik2006beyond} aims at preserving spatial information. Later works extend this approach with linear SPM \cite{yang2009linear} that selects the maximum SIFT features in a spatial region. Most of the early works on feature pooling have focused on max-pooling based on the max-like behavior of biological cortex signals \cite{serre2005object}. Maximum and average pooling studies in terms of information preservation by Boureau \textit{et al.} \cite{boureau2010theoretical} have suggested that max-pooling produces comparatively more representative results in low feature activation settings.

\textbf{Pooling in CNNs}. With the prevalence of learned feature approaches in various computer vision tasks, pooling methods have also been adapted to kernel-based operations. In CNNs, pooling has been mainly used to create condensed feature representations to reduce the model's computational requirements, and in turn to enable the creation of deeper architectures \cite{szegedy2015going}.

More recently, the preservation of relevant features during downsampling has taken a more prominent role. Initial approaches include stochastic pooling \cite{zeiler2013stochastic}, which uses a probabilistic weighted sampling of activations within a kernel region. Other pooling methods such as mixed pooling are based on a combination of maximum and average pooling, either probabilistically \cite{yu2014mixed} or through a combination of portions from each method \cite{lee2016generalizing}. Power Average ($L_{p}$) \cite{gulcehre2014learned} utilizes a learned parameter $p$ to determine the relative importance of average and maximum pooling. With $p=1$, sum pooling is used, while $p\rightarrow \infty$ corresponds to max-pooling.

Some approaches use grid-sampling. S3Pool \cite{zhai2017s3pool} randomly samples rows and columns of the original feature map to create the downsampled version. Methods can also employ learned weights such as in Detail Preserving Pooling (DPP, \cite{saeedan2018detail}) that uses average pooling while enhancing activations with above-average values. Local Importance Pooling (LIP, \cite{gao2019lip}) utilizes learned weights within a sub-network attention mechanism. A visual and mathematical overview of the operations performed by different pooling methods appears in Figure~\ref{fig:pooling_variants}.

The majority of the pooling work reported in the literature cannot be inverted for upsampling. Badrinarayanan \textit{et al.} \cite{badrinarayanan2017segnet} proposed an inversion of the maximum operation by tracking the in-kernel position of the selected maximum input while the other positions are populated by zero values in the upsampled output. This ensures that the original values are used, but the output is inherently sparse. Recently, Zhao and Snoek \cite{zhao2021liftpool} proposed LiftPool based on the use of four learnable sub-bands of the input. The produced output is composed as a mixture of the discovered sub-bands. They also propose an upsampling inversion of their approach (LiftUpPool). Both methods are based on sub-network structures that limit their usability as a computation and memory-efficient pooling technique.

Most of the aforementioned methods rely on combinations of maximum and average pooling, or the inclusion of sub-networks that prohibit low-compute and efficient downsampling. Instead of combining existing methods, our work is based on an adaptive exponential weighting approach to improve the retention of information and to better preserve details of the original signal. Our proposed method, adaPool, is inspired by Luce's choice of axiom \cite{luce1977choice}. We thus weigh kernel regions based on their relevance without being affected by the neighboring kernel vectors. This is in contrast to both average and maximum pooling. AdaPool uses two sets of pooling kernels. The first uses the channel-wise similarity of individual kernel vectors to their mean in order to determine their relevance. Similarities are calculated based on the Dice-S\o rensen coefficient. The second is based on softmax weighting to amplify feature activations of greater intensity \cite{stergiou2021refining}. Finally, outputs from both kernel operations are parametrically fused to a single volume. Parameters are specific to each kernel location thus making our approach regionally-adaptive.

A key property of adaPool is that gradients are calculated for each kernel vector during backpropagation. This improves the network connectivity. In addition, downsampled regions are less likely to exhibit a vanishing trend of activations, as observed by equal-contribution approaches such as average or sum pooling. We demonstrate how adaPool can adaptively capture details in Figure~\ref{fig:adapool_detail}, where the zoomed-in region displays a signature. AdaPool shows improvements in the clarity and recognizability of the letters and numbers.

\section{Methodology}
\label{sec:method}

In this section, we introduce the two processes (Sections~\ref{sec:method::sub:eDSC} and \ref{sec:method::sub:em}) that make up the final adaPool method (Section~\ref{sec:method::sub:adaPool}. We subsequently introduce the inverse adaUnPool method in Section~\ref{sec:method::sub:upsample}).

We start by introducing the basic operations of our pooling method. We define the local kernel region $\mathbf{R}$ as part of activation map $\mathbf{a}$ of size $C \! \times \! H \! \times \! W$, with $C$ channels, height $H$ and width $W$. For notation simplicity, we omit the channel dimension and assume that $\mathbf{R}$ is the set of relative position indices corresponding to the activations in the 2D spatial region of $k ! \times \! k$ (i.e., $|\mathbf{R}| = k^2$). We denote the pooling output as $\widetilde{\mathbf{a}}$ and the corresponding gradients as $\nabla\mathbf{\widetilde{a}}_i$, at the $i^{th}$ coordinate within region $\mathbf{R}$.

\subsection{Smooth approximated average pooling}
\label{sec:method::sub:eDSC}
Average pooling uses equal weights for all input vectors within a kernel region. The combined outputs are therefore strongly affected by outliers within the region. We argue that improvements in the calculation of the regional average can limit the effect of outlier values in both the creation of pooled volumes in the forward pass, as well as gradient calculations in the backward pass.

Inverse Distance Weighting (IDW) is widely applicable as a weighted average approach for multivariate interpolation \cite{akima1978method,franke1982scattered}. The assumption is that geometrically close observations exhibit a higher degree of resemblance than geometrically more distant ones. We extend IDW to kernel weighting for pooling by using the distance of each activation \textbf{a}$_{i}$, with coordinate index $i \in \mathbf{R}$, to the mean activation $\mathbf{\overline a}$ of $\mathbf{R}$. The resulting pooled region $\underset{IDW}{\widetilde{\mathbf{a}}}$ is formulated as: 

\begin{equation}
\label{eq:idpool}
    \underset{IDW}{\widetilde{\mathbf{a}}} =
    \begin{dcases}
    \sum\limits_{i \in \mathbf{R}}\frac{\underset{IDW}{w(\mathbf{\overline a},\mathbf{a}_{i})} \centerdot \mathbf{a}_{i}}{\sum\limits_{j \in \mathbf{R}} \underset{IDW}{w(\mathbf{\overline a},\mathbf{a}_{j})}},\; if \; d(\mathbf{\overline a},\mathbf{a}_{i}) \ne 0 \; \forall \; i \in \mathbf{R} \\
    \mathbf{a}_{i},\; if \; d(\mathbf{\overline a},\mathbf{a}_{i}) = 0 \; \exists \; i \in \mathbf{R}\\
    \end{dcases}
\end{equation}

The weights $\underset{IDW}{w(\cdot,\cdot})$ are based on the inverse of the distance $d(\cdot,\cdot)$ between each activation and the mean activation:

\begin{equation}
\label{eq:idweight}
    \underset{IDW}{w(\mathbf{\overline a},\mathbf{a}_{i})} =\frac{1}{d(\mathbf{\overline a},\mathbf{a}_{i})} \hspace{5em}
\end{equation}

Distance function $d(\cdot,\cdot)$ can be calculated by any geometric distance approach. Further details and limitations of IDWPool are discussed in Appendix~\ref{ap:A::IDW}.

As distance methods can produce artifacts when directly applied in input regions (see Appendix~\ref{ap:A::IDW}), the use of similarity measures is a better suited solution for the region-based nature of pooling. For the widely-used cosine similarity, an issue arises when the similarity between the two vectors is 1 even if one of the two vectors is infinitely large \cite{fernando2021anticipating}. Other dot-product methods for vector volumes such as the Dice-S\o rensen Coefficient (DSC) overcome this limitation by taking into account the vector lengths.

Given the IDW approach of Equation~\ref{eq:idpool}, zero-valued distances or coefficients will be assigned a zero weight. Therefore, our second extension is the use of the exponent ($e$) of the similarity between the activation vector and the average activations. This effectively makes the pooling method differentiable during backpropagation as at least a minimum gradient will be calculated for every location. It also reduces the possibility for the vanishing gradients problem to arise. Based on the introduction of the exponent of the similarity coefficient, we re-formulate Equation~\ref{eq:idpool} as:

\begin{equation}
\label{eq:edscwpool}
\underset{eDSC}{\widetilde{\mathbf{a}}} = \sum\limits_{i \in \mathbf{R}} \frac{e^{\underset{DSC}{w(\mathbf{\overline a},\mathbf{a}_{i})}} \centerdot \mathbf{a}_{i} 
}{\sum\limits_{j \in \mathbf{R}} e^{\underset{DSC}{w(\mathbf{\overline a},\mathbf{a}_{j})}}}
\end{equation}

It is important for downsamped volumes to preserve the informative features while reducing the spatial resolution of the input. The creation of volumes that do not fully capture the structural and feature appearances can have a negative impact on the performance. An example of such loss in detail can be seen in Figure~\ref{fig:adapool_detail}. Average pooling decreases the resolution of activations uniformly. Instead, using the exponent of the Dice-S\o rensen Coefficient ($e$DSCWPool) can improve on the activation preservation by exponentially weighting kernel values based on their similarity to their regional mean, while ensuring non-zero weights are assigned.

\subsection{Smooth approximated maximum pooling}
\label{sec:method::sub:em}

Complementary to the smooth approximated average within a kernel region, we discuss the formulation of downsampling based on the smooth approximated maximum which has been recently introduced as \textit{SoftPool} \cite{stergiou2021refining}. For clarity, and in line with the used terminology, we refer to SoftPool as exponential maximum pooling (\textit{e}MPool).

The motivation behind the use of the exponential maximum is influenced by the cortex neural simulations \cite{boureau2010theoretical,riesenhuber1999hierarchical} that downsample hand-coded features. The method is based on the natural exponent ($e$), which ensures that larger activations will have a greater effect on the final output while also ensuring that a minimum weight value is assigned to the lowest activations.


\begin{figure*}
\begin{center}
\includegraphics[width=.95\linewidth]{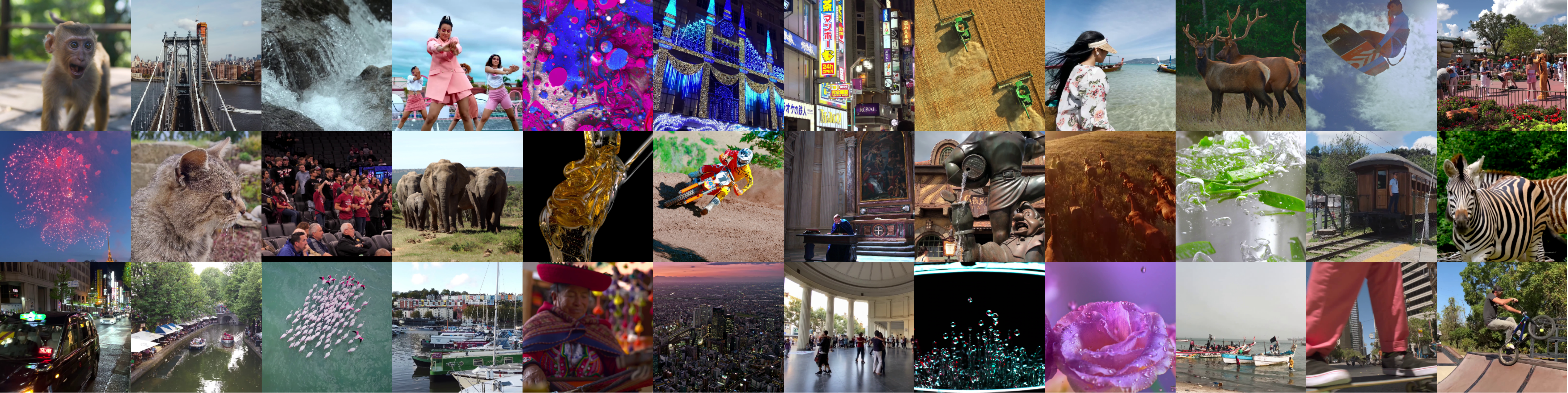}
\end{center}
   \caption{\textbf{Inter4K video frame samples.} These samples show the high resolution (UHD/4K) and variation in the frames. The videos are challenging for video processing due to rapid motions and movements, complex lighting, textures and object detail.}
\label{fig:inter4k_example}
\end{figure*}

The weights in exponential maximum pooling (\textit{e}MPool) are used as non-linear transforms based on the value of the corresponding activation. Higher-valued activations will become more dominant than lower-valued ones. As the majority of pooling operations are performed over high-dimensional feature spaces, highlighting the activations with greater effect is more balanced than the selection of the maximum activation alone. In the latter case, discarding the majority of the activations presents the risk of losing important information.

The output of \textit{e}MPool is produced through a summation of all weighted activations within the kernel region $\mathbf{R}$:

\begin{equation}
\label{eq:empool}
\underset{eM}{\widetilde{\mathbf{a}}} =\sum\limits_{i \in \mathbf{R}}\underset{eM}{w(\mathbf{a}_{i})}
\centerdot \mathbf{a}_{i}, \; \text{where} \; \underset{eM}{w(\mathbf{a}_{i})}=\frac{e^{\mathbf{a}_i}}{\sum\limits_{j \in \mathbf{R}}e^{\mathbf{a}_j}}
\end{equation}

\textit{e}MPool produces normalized results, similarly to \textit{e}DSCWPool. The results are based on a probability distribution that is proportional to the values of each activation with respect to the neighboring activations within the kernel region. 


\subsection{AdaPool: Adaptive exponential pooling}
\label{sec:method::sub:adaPool}

Based on their properties, \textit{e}DSCWPool uses the similarity of vectors $\mathbf{a}_{i}$ within the kernel region $\mathbf{R}$ to the mean activation $\overline{\mathbf{a}}$. \textit{e}MPool, however, uses the vectors in proportion to their values, with higher-valued activations being weighted more. From Figure~\ref{fig:adapool_detail}, neither of the two methods can be considered superior to the other. However, their properties can be complementary to discover the most informative features within the kernel region. With this observation, and in line with Lee \textit{et al.}'s introduction of average and maximum pooling fusion strategies \cite{lee2016generalizing}, we use a trainable weight mask $\boldsymbol\beta$ to create a combined volume of both smooth approximated average and smooth approximated maximum. Here, $\boldsymbol\beta$ is used to learn the proportion that will be used from each of the two methods within each kernel region $\mathbf{R}$. Introducing $\boldsymbol\beta$ as part of the network training process has the advantage of creating a generalized pooling strategy that relies on the combination of the properties of both \textit{e}MPool and \textit{e}DSCWPool. We formulate the method as a regionally-learned combination of the downsampled smooth approximated average  ($\underset{eDSC}{\widetilde{\mathbf{a}}}$) and the smooth approximated maximum ($\underset{eM}{\widetilde{\mathbf{a}}}$):

\begin{equation}
\label{eq:adapool}
\underset{ada}{\widetilde{\mathbf{a}}} \xlongequal{(3,4)} \underset{eDSC}{\widetilde{\mathbf{a}}} \centerdot \beta + \underset{eM}{\widetilde{\mathbf{a}}} \centerdot (1-\beta)
\end{equation}

where $\boldsymbol\beta \in \{0,...,1\}$ is a weight mask of the same size as the downsampled volume $\mathbf{\widetilde{a}}$ ($H' \! \times \! W'$). A visualization of adaPool appears in Figure~\ref{fig:adapool_formulation}. The gradients of $\boldsymbol{\beta}$ for backpropagation are calculated based on the chain rule as:

\begin{equation}
\label{eq:adapool_back}
\frac{\partial E}{\partial\beta} = \frac{\partial E}{\partial \underset{ada}{\widetilde{\mathbf{a}}}} \frac{\partial \underset{ada}{\widetilde{\mathbf{a}}}}{\partial \beta} = \frac{\partial E}{\partial \underset{ada}{\widetilde{\mathbf{a}}}} (\underset{i}{max} \; \mathbf{a}_{i} - \frac{1}{|R|} \sum_{i \in \textbf{R}}\mathbf{a}_i)
\end{equation}

\subsection{Upsampling using adaUnPool}
\label{sec:method::sub:upsample}

Pooling condenses regional information to a single output. The majority of the sub-sampling methods do not establish a bi-directional mapping between the sub-sampled and the original input, as most tasks do not require this link. However, tasks such as semantic segmentation \cite{jegou2017one, lin2017refinenet, wang2018understanding}, super-resolution \cite{chen2017deeplab, li2020mucan, lu2021masa, wang2020deep} or frame interpolation \cite{bao2019depth, liu2019deep, niklaus2018context, niklaus2020softmax} significantly benefit from it. As adaPool is differentiable and uses a minimum weight value assignment, the discovered weights can be used as prior knowledge during upsampling. We refer to this upsampling operation as \textit{adaUnPool}.

For a given pooled volume ($\mathbf{\widetilde a}$), we use the smooth approximated maximum ($\underset{eM}{w(\mathbf{a_{i}})}$) and smooth approximated average weights  ($\underset{eDSCW}{w(\mathbf{\overline{a}, a_{i}})}$) with learned weights mask $\boldsymbol \beta$. The final unpooled output ($\mathbf{a_{i}}$) for the $i$th kernel region ($i \in \mathbf{R}$) is computed as:

\begin{equation}
\label{eq:adaunpool}
\mathbf{a}_{i} = \beta \centerdot \frac{e^{\underset{DSC}{w}(\mathbf{\overline a},\mathbf{a}_{i})}}{\sum\limits_{j\in\mathbf{R}}e^{\underset{DSC}{w}(\mathbf{\overline a},\mathbf{a}_{j})}} \centerdot \mathcal{I}_{A}(\widetilde{\mathbf{a}}) + (1-\beta) \centerdot \underset{eM}{w(\textbf{a}_{i})} \centerdot \mathcal{I}_{A}(\widetilde{\mathbf{a}})
\end{equation}

where $\mathcal{I}_{A}(\cdot)$ interpolates by assigning the pooled volume ($\mathbf{\widetilde a}$) of the original kernel region at each position $i$. The method is used to inflate the volume from size $H' \! \times \! W'$ to $H \! \times \! W$.

\section{The Inter4K video dataset}
\label{sec:inter4k}
We introduce a novel high-resolution video dataset to benchmark upsampling methods. Inter4K is a collection of 1,000 ultra-high (4K) resolution clips with 60 frames per second (fps) sourced from YouTube. The dataset provides standardized video resolutions at ultra-high definition (UHD/4K), quad-high definition (QHD/2K), full-high definition (FHD/1080p), (standard) high definition (HD/720p), one quarter of full HD (qHD/520p) and one ninth of a full HD (nHD/360p). Available frame rates for each resolution include 60, 50, 30, 24 and 15 fps. Based on this standardization, both super-resolution and frame interpolation tests can be performed for different scaling sizes ($\times 2$, $\times 3$, and $\times 4$). In our experiments, we use Inter4K to address both tasks of frame upsampling and interpolation.

\begin{figure}[!tbp]
\RawFloats
  \centering
  \begin{minipage}[b]{0.46\textwidth}
    \includegraphics[width=\textwidth]{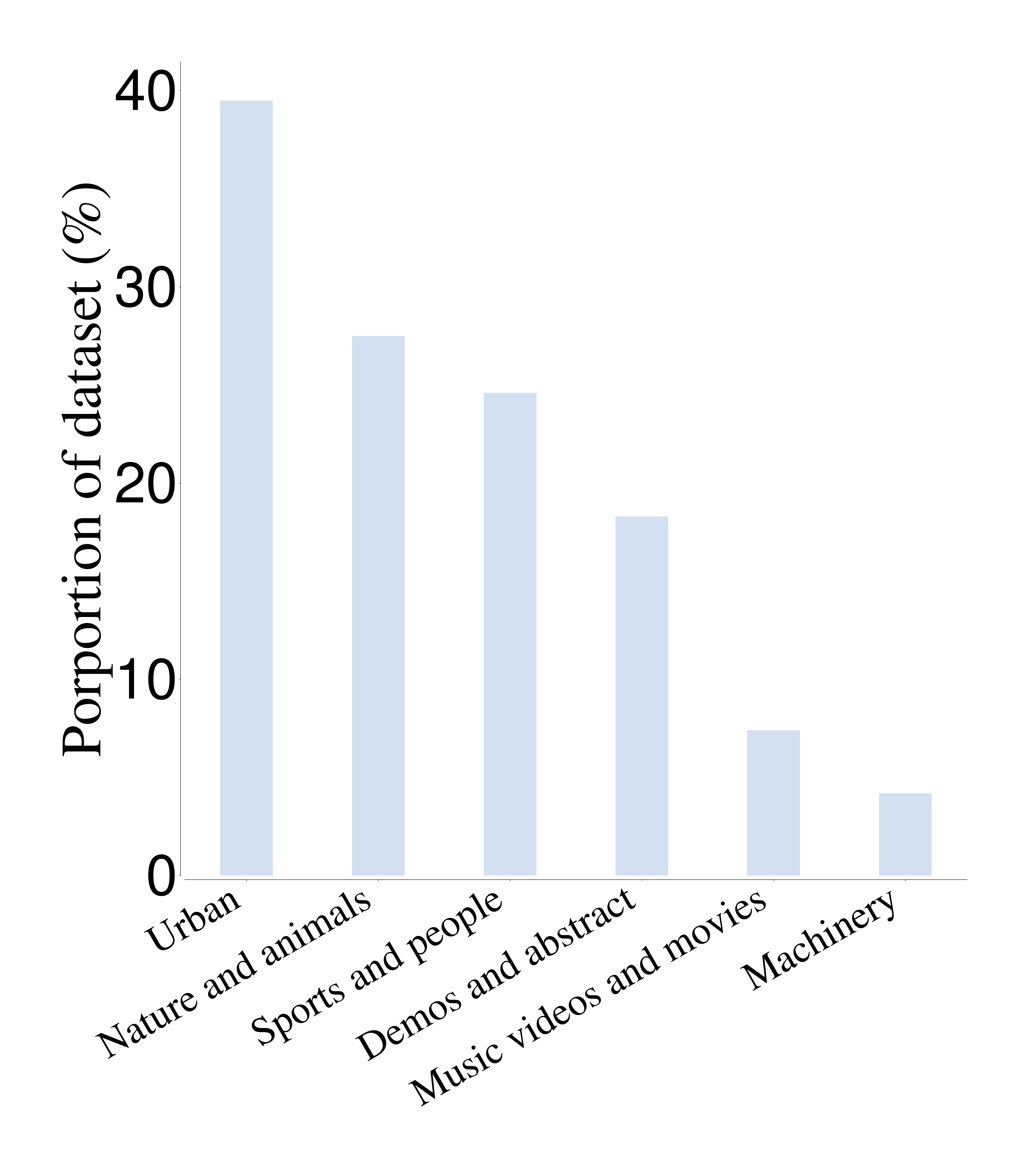}
    \caption{\textbf{Inter4K category proportions}. Categories are selected based on broad concepts of the videos.}
    \label{fig:Inter4k_ontology}
  \end{minipage}
  \hfill
  \begin{minipage}[b]{0.46\textwidth}
    \includegraphics[width=\textwidth]{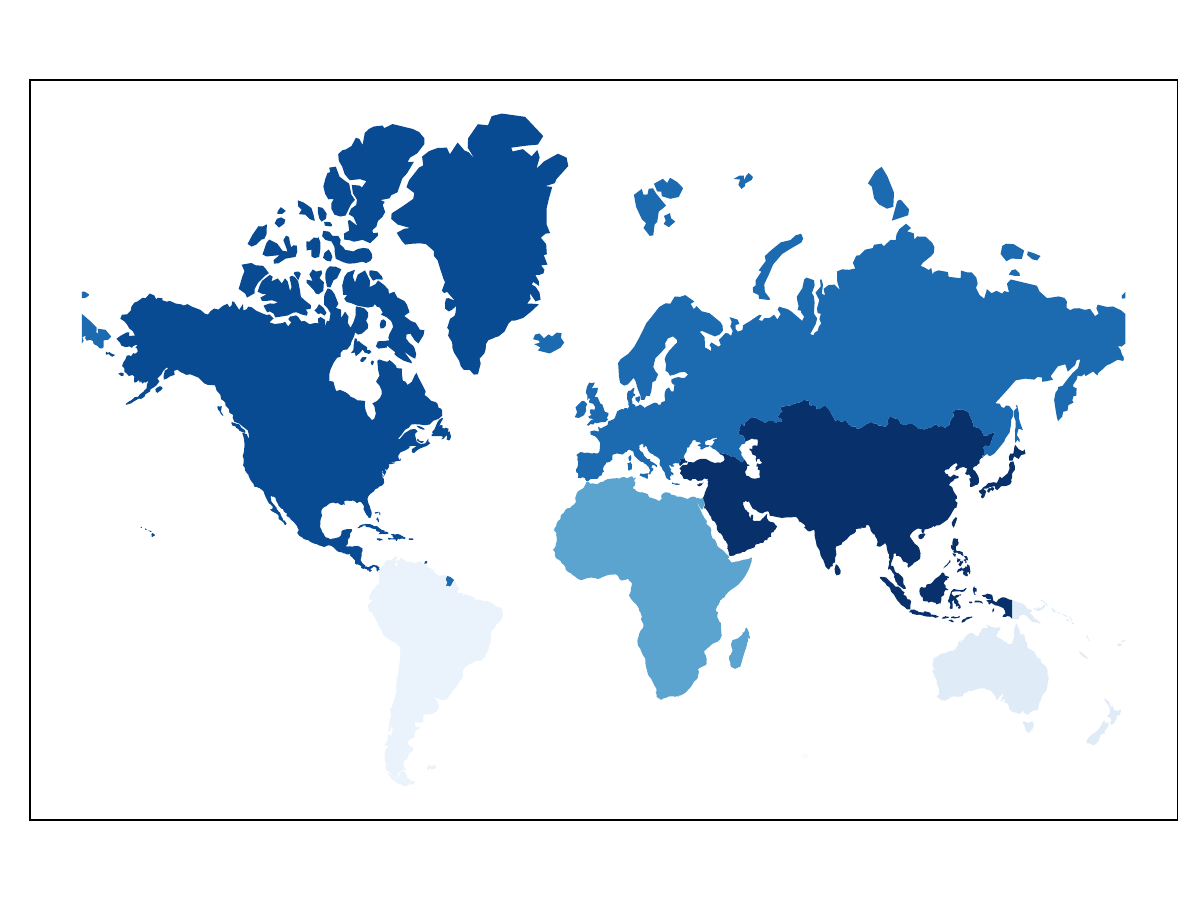}
    \vspace{1.25em}
    \caption{\textbf{Inter4K video locations by continent}. Darker colors correspond to a larger number of videos.}
    \label{fig:Inter4k_map}
  \end{minipage}
  \vspace{-1pt}
\end{figure}

In contrast to other datasets used for video super-resolution and interpolation \cite{baker2011database,liu2013bayesian,mercat2020uvg,soomro2012ucf101,takeda2009super,yi2019progressive,xue2019video}, Inter4K provides standardized UHD resolution at 60 fps for all videos. The dataset is divided into 800 videos for training, 100 videos for validation, and 100 videos for testing. Videos are of 5-second length (examples are shown in Figure~\ref{fig:inter4k_example}) and include diverse scenes based on equipment used (e.g., professional 4K cameras, mobile phones), lighting conditions, static and moving cameras, and variations in movements, actions, and objects. We include a summary of the videos in Inter4K based on six main categories as presented in Figure~\ref{fig:Inter4k_ontology}. Categories are chosen given the primary focus of the video. The main four categories that correspond to ~90\% of the videos include \textit{Urban} environments (e.g. buildings, streets, or vehicles), \textit{Nature and animals}, \textit{Sports and people} depicting human activities and actions, and \textit{Demos and abstract} with demo videos for video resolution and frame rates, or videos with computer-generated abstract shapes. The last two categories are less prevalent in the dataset either due to copyright restrictions (\textit{Music videos and movies}) or scarcity of videos (\textit{Machinery}). In Figure~\ref{fig:Inter4k_map} we present a visualization of the locations of 632 out of the 1,000 videos. These locations were found based on available geo-tags, video titles, and keywords or depictions of identifiable landmarks in the video. Both Figures~\ref{fig:Inter4k_ontology} and \ref{fig:Inter4k_map} demonstrate the diversity of Inter4K in terms of video content and the locations where the videos were shot.

\section{Main Results} \label{sec:experiments}

We initially evaluate the information loss caused by downsampling with various pooling methods. We compare the downsampled and original images using standard similarity measures (Section~\ref{sec:experiments::similarity}). In addition, we examine the computational overhead of each pooling method (Section~\ref{sec:experiments::latency}).

We proceed by testing the performance of widely-used CNN architectures on ImageNet1K when we substitute the network's original pooling layers by \textit{e}MPool, \textit{e}DSCWPool and adaPool (Section~\ref{sec:experiments::imagenet}). We also provide comparisons between different pooling methods (Section~\ref{sec:experiments::pooling_methods}).

We present our results for object detection (Section~\ref{sec:experiments::detection}) on MS COCO \cite{lin2014microsoft} with RetinaNet~\cite{lin2017focal} and Mask R-CNN~\cite{he2017mask} using several backbones. We additionally experiment on spatio-temporal data by focusing on action recognition in video (Section~\ref{sec:experiments::video}).

Lastly, we present our results on image super-resolution, frame interpolation, and their combination (Section~\ref{sec:experiments::superres_iner}).

\subsection{Experimental settings}
\label{sec:experiments::settings}

\textbf{Datasets}. For our image-based experiments, we use seven different datasets for quantitative evaluation of the downsampled image quality, image classification, object detection, and image super-resolution. For the assessment of image quality and similarity, we use the high-resolution DIV2K \cite{agustsson2017ntire}, Urban100 \cite{huang2015single}, Manga109 \cite{matsui2017sketch}, and Flicker2K \cite{agustsson2017ntire} datasets. For image classification we use ImageNet1K \cite{ILSVRC15}, and MS COCO \cite{lin2014microsoft} for image object detection. For image super-resolution we employ the Urban100, Manga109, and B100 \cite{martin2001database} datasets.
For our video-based experiments, we employ six datasets. For action recognition, we use the large-scale HACS \cite{zhao2019hacs} and Kinetics-700 \cite{carreira2019short} datasets, as well as the smaller UCF-101 \cite{soomro2012ucf101} dataset. For frame interpolation, we use Vimeo90K \cite{xue2019video} and Middlebury~\cite{baker2011database} video processing datasets, as well as our newly introduced Inter4K dataset, which is also used for the combined task of frame interpolation and super-resolution.

\textbf{Classification training scheme}. For image classification, we use a random spatial region crop of size $294 \times 294$, which is then resized to $224 \times 224$. The initial learning rate across our experiments is set to 0.1 with an SGD optimizer. We train for a total of 100 epochs with a step-wise learning rate reduction every 40 epochs. For higher numbers of epochs, no further improvements were observed. The batch size is set to 256.

For our video action recognition tests, we use a multigrid training scheme \cite{wu2020multigrid}, with frame sizes between 4--16 and frame crops of 90--256 depending on the cycle. The average video inputs are of size $8 \times 160 \times 160$, while the batch sizes are between 64 and 2048. The size for each of the batches is counter-equal to the input size in every step in order to optimize memory use. We use the same learning rate, optimizer, learning rate schedule, and maximum number of epochs as in the image-based experiments.

\textbf{Object detection details}. We first rescale the images to ensure that the smallest side has a minimum size of at least 800 pixels \cite{lin2017feature,he2017mask}. If after rescaling the largest side is larger than 1024 pixels, we resize the entire image so that the largest side becomes 1024 pixels. Our rescaling and resizing preserves the aspect ratio of the images. We use the pre-trained networks from the image classification task as backbones. The learning rate is set to $1e\!-\!5$ and we use an SGD optimizer with 0.9 momentum.

\begin{table*}[t]
\setlength\tabcolsep{2.4pt}
\caption{\textbf{Quantitative results on benchmark high-res datasets}. Best results for each setting are denoted in \textbf{bold}.} 
\label{tab:times_ssim_psnr}
\centering
\resizebox{\textwidth}{!}{%
\begin{tabular}{cll|ccc|ccc|ccc|ccc|ccc|ccc|ccc|ccc|ccc}
\hline
&
\multicolumn{2}{c|}{\multirow{3}{*}{Pooling method}} &
\multicolumn{9}{c|}{DIV2K \cite{agustsson2017ntire}} &
\multicolumn{9}{c|}{Urban100 \cite{huang2015single}} & 
\multicolumn{9}{c}{Manga109 \cite{matsui2017sketch}} \tstrut \\[.3em]
& & &
\multicolumn{3}{c|}{$k=2$} &
\multicolumn{3}{c|}{$k=3$} &
\multicolumn{3}{c|}{$k=5$} &
\multicolumn{3}{c|}{$k=2$} &
\multicolumn{3}{c|}{$k=3$} &
\multicolumn{3}{c|}{$k=5$} &
\multicolumn{3}{c|}{$k=2$} &
\multicolumn{3}{c|}{$k=3$} &
\multicolumn{3}{c}{$k=5$} \\[.2em]\cline{4-30}
& 
&
& SSIM & PSNR & LPIPS
& SSIM & PSNR & LPIPS
& SSIM & PSNR & LPIPS

& SSIM & PSNR & LPIPS
& SSIM & PSNR & LPIPS 
& SSIM & PSNR & LPIPS 

& SSIM & PSNR & LPIPS
& SSIM & PSNR & LPIPS 
& SSIM & PSNR & LPIPS \tstrut \bstrut \\
\hline
\parbox[t]{2mm}{\multirow{3}{*}{\rotatebox[origin=c]{90}{}}} 

& \multicolumn{2}{l|}{Avg}  
& 0.714 & 51.247 & 0.204
& 0.578 & 44.704 & 0.325
& 0.417 & 29.223 & 0.494
& 0.691 & 50.380 & 0.212
& 0.563 & 41.745 & 0.352
& 0.372 & 28.270 & 0.416
& 0.695 & 54.326 & 0.189
& 0.582 & 43.657 & 0.277
& 0.396 & 29.862 & 0.374
\tstrut \\[.15em]

& \multicolumn{2}{l|}{Max} 
& 0.685 & 49.826 & 0.229
& 0.370 & 41.944 & 0.367
& 0.358 & 22.041 & 0.524
& 0.662 & 48.266 & 0.252
& 0.528 & 40.709 & 0.405
& 0.330 & 20.654 & 0.476
& 0.671 & 50.085 & 0.210
& 0.544 & 41.128 & 0.324
& 0.324 & 22.307 & 0.413
\\[.15em]

& \multicolumn{2}{l|}{Pow-avg}
& 0.419 & 35.587 & 0.323
& 0.286 & 26.329 & 0.500
& 0.178 & 16.567 & 0.657
& 0.312 & 31.911 & 0.313
& 0.219 & 24.698 & 0.512
& 0.124 & 15.659 & 0.591
& 0.381 & 29.248 & 0.313
& 0.276 & 18.874 & 0.460
& 0.160 & 9.266 & 0.553
\\[.15em]

& \multicolumn{2}{l|}{Sum}
& 0.408 & 35.153 & 0.315
& 0.268 & 26.172 & 0.489
& 0.193 & 17.315 & 0.634
& 0.301 & 31.657 & 0.308
& 0.208 & 24.735 & 0.511
& 0.123 & 15.243 & 0.602
& 0.374 & 30.169 & 0.321
& 0.271 & 20.150 & 0.456
& 0.168 & 13.081 & 0.549
\bstrut \\\cline{1-30}

\parbox[t]{2mm}{\multirow{4}{*}{\rotatebox[origin=c]{90}{Trainable}}} 
& \multicolumn{2}{l|}{$L_{p}$ \cite{gulcehre2014learned}}
& 0.686 & 49.912 & 0.213
& 0.542 & 43.083 & 0.328
& 0.347 & 25.139 & 0.502
& 0.676 & 48.508 & 0.224
& 0.534 & 39.986 & 0.365
& 0.326 & 26.365 & 0.453
& 0.675 & 51.721 & 0.206
& 0.561 & 41.824 & 0.292
& 0.367 & 27.469 & 0.407
\tstrut \\[.15em]

& \multicolumn{2}{l|}{Gate \cite{lee2016generalizing}}
& 0.689 & 50.104 & 0.211
& 0.560 & 43.437 & 0.326
& 0.353 & 25.672 & 0.497
& 0.675 & 49.769 & 0.212
& 0.537 & 40.422 & 0.363
& 0.328 & 26.731 & 0.433
& 0.679 & 51.980 & 0.198
& 0.569 & 42.127 & 0.283
& 0.374 & 27.754 & 0.393
\\[.15em]

& \multicolumn{2}{l|}{DPP \cite{saeedan2018detail}}  
& 0.702 & 50.598 & 0.206
& 0.562 & 44.076 & 0.324
& 0.396 & 27.421 & 0.498
& 0.684 & 49.947 & 0.213
& 0.551 & 40.813 & 0.357
& 0.365 & 27.136 & 0.425
& 0.691 & 52.646 & 0.191
& 0.573 & 42.794 & 0.294
& 0.386 & 28.598 & 0.386
\\[.15em]

& \multicolumn{2}{l|}{LIP \cite{gao2019lip}}
& 0.711 & 50.831 & 0.203
& 0.559 & 44.432 & 0.323
& 0.401 & 28.285 & 0.492
& 0.689 & 50.266 & 0.212
& 0.558 & 41.159 & 0.354
& 0.370 & 27.849 & 0.415
& 0.689 & 53.537 & 0.185
& 0.579 & 43.018 & 0.273
& 0.391 & 29.331 & 0.373
\bstrut \\\cline{1-30}

\parbox[t]{2mm}{\multirow{2}{*}{\rotatebox[origin=c]{90}{Stoch.}}} 
& \multicolumn{2}{l|}{Stochastic \cite{zeiler2013stochastic}} 
& 0.631 & 45.362 & 0.321
& 0.479 & 39.895 & 0.497
& 0.295 & 21.314 & 0.609
& 0.616 & 44.342 & 0.285
& 0.463 & 37.223 & 0.476
& 0.286 & 19.358 & 0.561
& 0.583 & 46.274 & 0.316
& 0.427 & 39.259 & 0.433
& 0.255 & 22.953 & 0.521
\tstrut \\[.3em]

& \multicolumn{2}{l|}{S3 \cite{zhai2017s3pool}} 
& 0.609 & 44.760 & 0.318
& 0.454 & 39.326 & 0.486
& 0.280 & 20.773 & 0.615
& 0.608 & 44.239 & 0.276
& 0.459 & 36.965 & 0.463
& 0.272 & 19.645 & 0.548
& 0.576 & 46.613 & 0.305
& 0.426 & 39.866 & 0.427
& 0.232 & 23.242 & 0.521
\bstrut \\

\cline{1-30}

\parbox[t]{2mm}{\multirow{5}{*}{\rotatebox[origin=c]{90}{IDW \textbf{(Ours)}}}}
& \multicolumn{2}{l|}{$L_1$}  
& 0.724 & 51.415 & 0.218
& 0.596 & 44.739 & 0.346
& 0.418 & 29.576 & 0.511
& 0.696 & 50.723 & 0.218
& 0.573 & 41.796 & 0.367
& 0.366 & 28.205 & 0.437
& 0.712 & 54.614 & 0.195
& 0.574 & 43.756 & 0.283
& 0.395 & 30.224 & 0.389
\tstrut \\[.15em]
& \multicolumn{2}{l|}{$L_2$}
& 0.726 & 51.421 & 0.217
& 0.601 & 44.753 & 0.346
& 0.421 & 29.581 & 0.504
& 0.698 & 50.731 & 0.217
& 0.575 & 41.794 & 0.368
& 0.372 & 28.211 & 0.433
& 0.711 & 54.617 & 0.191
& 0.579 & 43.762 & 0.281
& 0.404 & 30.256 & 0.379
\\[.15em]
& \multirow{3}{*}{\rotatebox[origin=c]{90}{\footnotesize Huber \cite{huber1992robust}}} & $\delta= 1/4$
& 0.728 & 51.465 & 0.215
& 0.611 & 44.813 & 0.336
& 0.427 & 29.736 & 0.502
& 0.702 & 50.734 & 0.216
& 0.579 & 41.862 & 0.364
& 0.383 & 28.576 & 0.429
& 0.714 & 54.620 & 0.192
& 0.584 & 43.897 & 0.278
& 0.412 & 30.445 & 0.378
\\[.15em]
& & $\delta= 1/2$
& 0.730 & 51.487 & 0.214
& 0.617 & 44.924 & 0.327
& 0.429 & 29.861 & 0.497
& 0.710 & 50.742 & 0.216
& 0.581 & 41.916 & 0.359
& 0.389 & 28.674 & 0.431
& 0.721 & 54.637 & 0.190
& 0.588 & 43.936 & 0.278
& 0.421 & 30.429 & 0.376
\\[.15em]
& & $\delta = 3/4$
& 0.727 & 51.459 & 0.215
& 0.606 & 44.846 & 0.330
& 0.421 & 29.728 & 0.499
& 0.705 & 50.737 & 0.215
& 0.576 & 41.874 & 0.358
& 0.384 & 28.612 & 0.428
& 0.716 & 54.626 & 0.188
& 0.581 & 43.885 & 0.277
& 0.409 & 30.432 & 0.375
\bstrut \\

\cline{1-30}

\parbox[t]{2mm}{\multirow{6}{*}{\rotatebox[origin=c]{90}{exponential \textbf{(Ours) }}}}
& \multicolumn{2}{l|}{\textit{e}M}
& 0.729 & 51.436 & 0.204
& 0.594 & 44.747 & 0.339
& 0.421 & 29.583 & 0.498
& 0.694 & 50.687 & 0.211
& 0.578 & 41.851 & 0.352
& 0.394 & 28.326 & 0.417
& 0.704 & 54.563 & 0.189
& 0.586 & 43.782 & 0.276
& 0.403 & 30.114 & 0.373
\tstrut \\[.15em]
& \multicolumn{2}{l|}{\textit{e}DSCW} 
& 0.732 & 51.470 & 0.203
& 0.619 & 45.324 & 0.324
& 0.430 & 30.247 & 0.498
& 0.706 & 50.734 & 0.211
& 0.588 & 42.173 & 0.351
& 0.412 & 29.796 & 0.413
& 0.715 & 54.633 & 0.173
& 0.593 & 44.185 & 0.268
& 0.417 & 30.967 & 0.371
\\[.15em]
& \multicolumn{2}{l|}{adapt. $\boldsymbol{\beta} \! = \! 1/4$}  
& 0.730 & 51.523 & 0.199
& 0.605 & 44.979 & 0.321
& 0.429 & 29.956 & 0.492
& 0.712 & 50.733 & 0.199
& 0.582 & 41.970 & 0.345
& 0.397 & 28.861 & 0.407
& 0.709 & 54.625 & 0.168
& 0.589 & 43.994 & 0.265
& 0.412 & 30.843 & 0.368
\\[.15em]
& \multicolumn{2}{l|}{adapt. $\boldsymbol{\beta} \! = \! 1/2$}  
& 0.736 & 52.186 & 0.198
& 0.614 & 45.857 & 0.321
& 0.432 & 30.881 & 0.491
& 0.754 & 51.376 & 0.197
& 0.594 & 42.462 & 0.345
& 0.410 & 29.874 & 0.406
& 0.716 & 54.646 & 0.167
& 0.598 & 44.576 & 0.264
& 0.421 & 31.392 & 0.364
\\[.15em]
& \multicolumn{2}{l|}{adapt. $\boldsymbol{\beta} \! = \! 3/4$} 
& 0.742 & 53.341 & 0.194
& 0.618 & 46.173 & 0.319
& 0.438 & 31.764 & 0.489
& 0.741 & 51.132 & 0.196
& 0.605 & 42.996 & 0.341
& 0.418 & 30.163 & 0.405
& 0.731 & 55.284 & 0.164
& 0.602 & 45.344 & 0.260
& 0.427 & 31.638 & 0.366
\\[.15em]
& \multicolumn{2}{l|}{adapt. $\nabla \boldsymbol{\beta}$}  
& \textbf{0.778} & \textbf{53.769} & \textbf{0.184} 
& \textbf{0.624} & \textbf{46.892} & \textbf{0.305} 
& \textbf{0.443} & \textbf{32.216} & \textbf{0.483} 
& \textbf{0.769} & \textbf{52.438} & \textbf{0.188} 
& \textbf{0.614} & \textbf{43.637} & \textbf{0.337} 
& \textbf{0.425} & \textbf{30.845} & \textbf{0.396} 
& \textbf{0.747} & \textbf{55.890} & \textbf{0.156} 
& \textbf{0.609} & \textbf{46.253} & \textbf{0.254} 
& \textbf{0.436} & \textbf{32.794} & \textbf{0.355} 
\\[.15em]
\end{tabular}
}
\end{table*}

\begin{table}[t]
\caption{\textbf{Latency and pixel similarity}. Latency for the forward and backward pass on both CPU and GPU is averaged over all images in ImageNet1K. Pixel similarity reported on Flicker2K. Best results in \textbf{bold}.
}
\label{tab:compute_flicker}
\centering
\resizebox{\linewidth}{!}{%
\setlength\tabcolsep{2pt}
\begin{tabular}{ll|l|l|ccc|ccc|ccc}
\hline
\multicolumn{2}{l|}{\multirow{3}{*}{Pooling}} &
\multicolumn{1}{c|}{CPU} &
\multicolumn{1}{c|}{CUDA} &
\multicolumn{9}{c}{Flicker2K \cite{agustsson2017ntire}} \tstrut \\
 & & \multicolumn{1}{c|}{(ms)} & \multicolumn{1}{c|}{(ms)} &
\multicolumn{3}{c|}{$k=2$} &
\multicolumn{3}{c|}{$k=3$} &
\multicolumn{3}{c}{$k=5$} \bstrut \\\cline{3-13}
& & ($\downarrow \! \text{F} / \! \uparrow \! \text{B})$ & ($\downarrow \! \text{F} / \! \uparrow \! \text{B}$) 
& SSIM & PSNR & LPIPS
& SSIM & PSNR & LPIPS
& SSIM & PSNR & LPIPS \tstrut \bstrut \\
\hline

\multicolumn{2}{l|}{Avg} & $9 \, \! / \, \! 49$ & $14 \, \! / \, \! 76$ 
& 0.709 & 51.786 & 0.2138
& 0.572 & 44.246 & 0.3289
& 0.408 & 28.957 & 0.5020 \tstrut \\[.15em]

\multicolumn{2}{l|}{Max} & $91 \, \! / \, \! 152$ & $195 \, \! / \, \! 267$ 
& 0.674 & 47.613 & 0.2367
& 0.385 & 40.735 & 0.3781
& 0.329 & 21.368 & 0.5372 \\[.15em]

\multicolumn{2}{l|}{Pow-avg} & $74 \, \! / \, \! 329$ & $120 \, \! / \, \! 433$ 
& 0.392 & 34.319 & 0.3214
& 0.271 & 26.820 & 0.5346
& 0.163 & 15.453 & 0.6608 \\[.15em]

\multicolumn{2}{l|}{Sum} & $26 \, \! / \, \! 163$ & $79 \, \! / \, \! 323$ 
& 0.386 & 34.173 & 0.3276
& 0.265 & 26.259 & 0.5212
& 0.161 & 15.218 & 0.6573 \bstrut \\

\cline{1-13}

\multicolumn{2}{l|}{$L_{p}$ \cite{gulcehre2014learned}} & $116 \, \! / \, \! 338$ & $214 \, \! / \, \! 422$
& 0.683 & 48.617 & 0.2269
& 0.437 & 42.079 & 0.3572
& 0.341 & 24.432 & 0.5326 \tstrut \\[.15em]

\multicolumn{2}{l|}{Gate \cite{lee2016generalizing}} & $245 \, \! / \, \! 339$ & $327 \, \! / \, \! 540$ 
& 0.687 & 49.314 & 0.2241
& 0.449 & 42.722 & 0.3453
& 0.358 & 25.687 & 0.5245 \\[.15em]

\multicolumn{2}{l|}{DPP \cite{saeedan2018detail}} & $427 \, \! / \, \! 860$ & $634 \, \! / \, \! 1228$
& 0.691 & 50.586 & 0.2155
& 0.534 & 43.608 & 0.3341
& 0.385 & 27.430 & 0.5137 \\[.15em]

\multicolumn{2}{l|}{LIP \cite{gao2019lip}} & $134 \, \! / \, \! 257$ & $258 \, \! / \, \! 362$  
& 0.696 & 50.947 & 0.2140
& 0.548 & 43.882 & 0.3292
& 0.390 & 28.134 & 0.5034 \bstrut \\

\cline{1-13}

\multicolumn{2}{l|}{Stoch. \cite{zeiler2013stochastic}} & $162 \, \! / \, \! 341$ & $219 \, \! / \, \! 485$ 
& 0.625 & 46.714 & 0.3416
& 0.474 & 38.365 & 0.4876
& 0.264 & 21.428 & 0.5724 \tstrut \\[.3em]

\multicolumn{2}{l|}{S3 \cite{zhai2017s3pool}} & $233 \, \! / \, \! 410$ & $345 \, \! / \, \! 486$  
& 0.611 & 46.547 & 0.3205
& 0.476 & 37.706 & 0.4531
& 0.252 & 21.363 & 0.5640 \bstrut \\

\cline{1-13}

\multicolumn{2}{l|}{$L_{1}$} & N/A & $48 \, \! / \, \! 227$  
& 0.712 & 51.985 & 0.2283
& 0.574 & 44.814 & 0.3365
& 0.411 & 29.246 & 0.5208 \tstrut \\[.3em]

\multicolumn{2}{l|}{$L_{2}$} & N/A & $49 \, \! / \, \! 231$  
& 0.714 & 51.592 & 0.2276
& 0.576 & 44.832 & 0.3342
& 0.416 & 29.251 & 0.5188 \\[.2em]

\multirow{3}{*}{\rotatebox[origin=c]{90}{\footnotesize Huber \cite{huber1992robust}}} & $\delta \! = \! 1/4$ & N/A & $51 \, \! / \, \! 234$  
& 0.726 & 51.879 & 0.2265
& 0.583 & 44.916 & 0.3314
& 0.421 & 29.458 & 0.5162 \\[.3em]

& $\delta \! = \! 1/2$ & N/A & $51 \, \! / \, \! 234$  
& 0.727 & 52.053 & 0.2214
& 0.593 & 45.231 & 0.3299
& 0.424 & 29.635 & 0.5134 \\[.3em]

& $\delta \! = \! 3/4$ & N/A & $51 \, \! / \, \! 234$  
& 0.723 & 51.912 & 0.2218
& 0.584 & 45.105 & 0.3312
& 0.418 & 29.476 & 0.5127 \bstrut \\

\cline{1-13}

\multicolumn{2}{l|}{\textit{e}M} & $31 \, \! / \, \! 156$ & $56 \, \! / \, \! 234$ 
& 0.721 & 52.356 & 0.2143
& 0.587 & 44.893 & 0.3297
& 0.416 & 29.341 & 0.5026 \tstrut \\[.15em]

\multicolumn{2}{l|}{\textit{e}DSCW} & N/A & $52 \, \! / \, \! 249$ 
& 0.743 & 53.293 & 0.2124
& 0.615 & 45.274 & 0.3275
& 0.421 & 29.957 & 0.4986 \\[.15em]

\multicolumn{2}{l|}{adaptive} & $N/A$ & $119 \, \! / \, \! 490$ 
& \textbf{0.766} & \textbf{54.608} & \textbf{0.2045}
& \textbf{0.629} & \textbf{47.139} & \textbf{0.3089}
& \textbf{0.434} & \textbf{31.883} & \textbf{0.4831} \\[.15em]

\end{tabular}
}
\end{table}

\subsection{Downsampling similarity}
\label{sec:experiments::similarity}

In the first set of tests, we evaluate the information loss when using our proposed methods for downsampling. The comparisons focus on the similarity of the original inputs and downsampled outputs. Three widely used pooling kernel sizes are employed ($k \! = \! \{2,3,5\}$). We use three standardized evaluation metrics \cite{wang2004image,zhang2018unreasonable}:

\noindent
\textbf{Structural Similarity Index Measure (SSIM)} is calculated as the difference of two images in terms of their luminance, contrast, and a structural term. Larger SSIM values correspond to larger structural similarities.

\noindent
\textbf{Peak Signal-to-Noise Ratio (PSNR)} is a quantification of the produced image's compression quality. PSNR takes into account the inverse of the Mean Squared Error (MSE) of two images' channels. Higher PSNR values translate to smaller channel-wise distances between the two images.

\noindent
\textbf{Learned Perceptual Image Patch Similarity (LPIPS)} is a similarity measurement between patches from two images. LPIPS compares the distances of features from the two images extracted by a deep learning backbone. Lower LPIPS values correspond to higher similarity between images.

In Tables~\ref{tab:times_ssim_psnr} and \ref{tab:compute_flicker}, we present the SSIM, PSNR, and LPIPS values averaged over all images in DIV2K \cite{agustsson2017ntire}, Urban100 \cite{huang2015single}, Manga109 \cite{matsui2017sketch}, and Flicker2K \cite{agustsson2017ntire} datasets, for different kernel sizes. IDW-based distance methods outperform non-trainable and stochastic methods. The randomized policy of stochastic methods does not seem to allow to fully capture details. Additionally, the use of exponential weighting to our IDW-based methods yields clear improvements. Both \textit{e}MPool and \textit{e}DSCWPool are top-performing across kernel sizes and datasets, demonstrating the benefits of exponential approximation methods for image downsampling. Finally, the combination of the two exponential methods into adaPool consistently achieves the best overall performance when the fusion parameter $\boldsymbol{\beta}$ is learned.

\subsection{Latency and memory use} 
\label{sec:experiments::latency}

Costs in terms of the memory and latency required by pooling operations are largely overlooked in literature as single operations have minor latency times and memory consumption. However, given potentially limited available resources, and the fact that operations are executed thousands of times per epoch, we advocate an evaluation of the running times and memory use. Slow or memory-intensive operations can have a detrimental effect on the performance and may become potential computational bottlenecks.

Computation overheads are reported in Table~\ref{tab:compute_flicker} based on the inference over CPU and GPU (CUDA) for forward ($\downarrow F$) and backward ($\uparrow B$) passes over each operation. We observe that our implementations achieve reasonable inference times on CUDA despite the additional computations in comparison to methods such as average, maximum, power average or sum pooling.

\begin{table*}[t]
\caption{\textbf{Pairwise comparisons of top-1 and top-5 accuracies} on ImageNet1K \cite{ILSVRC15} between original networks and their counterparts with pooling replaced by \textit{e}mPool, \textit{e}DSCWPool and adaPool. All networks have been trained from scratch. Best results in \textbf{bold}. More details for the parameters and FLOPs are provided in Appendix~\ref{ap:A:compute}.}
\label{tab:ImageNet_scratch}
\centering
\resizebox{.95\linewidth}{!}{%
\begin{tabular}{l|cc|cc|cc|cc|cc}
\hline
\multirow{2}{*}{Model} & 
Params & \multirow{2}{*}{GFLOPs} &
\multicolumn{2}{c|}{Original (Baseline)} & \multicolumn{2}{c|}{\textit{e}MPool} & \multicolumn{2}{c|}{\textit{e}DSCWPool}  & 
\multicolumn{2}{c}{adaPool} \tstrut \\
& (M) & & top-1 & top-5 & top-1 & top-5 & top-1 & top-5 & top-1 & top-5 \bstrut \\
\hline
ResNet-18 & 11.7 & 1.83 & 
69.76 & 
89.08 &
71.27 (\textcolor{applegreen}{+1.51}) & 
90.16 (\textcolor{applegreen}{+1.08}) &
70.79 (\textcolor{applegreen}{+1.03}) &
89.96 (\textcolor{applegreen}{+0.88}) &
\textbf{71.78} (\textcolor{applegreen}{+2.02}) & 
\textbf{90.65} (\textcolor{applegreen}{+1.57}) \tstrut \\[0.1em]

ResNet-34 & 21.8 & 3.68 & 
73.30 & 
91.42 &
74.67 (\textcolor{applegreen}{+1.37}) & 
92.30 (\textcolor{applegreen}{+0.88}) &
74.36 (\textcolor{applegreen}{+1.06}) &
92.15 (\textcolor{applegreen}{+0.73}) &
\textbf{75.43} (\textcolor{applegreen}{+2.13}) & 
\textbf{92.87} (\textcolor{applegreen}{+1.45}) \\[0.1em]

ResNet-50 & 25.6 & 4.14 & 
76.15 & 
92.87 &
77.35 (\textcolor{applegreen}{+1.17}) & 
93.63 (\textcolor{applegreen}{+0.76}) &
77.38 (\textcolor{applegreen}{+1.23}) &
93.90 (\textcolor{applegreen}{+1.03}) &
\textbf{78.42} (\textcolor{applegreen}{+2.27}) & 
\textbf{94.16} (\textcolor{applegreen}{+1.29})\\[0.1em]

ResNet-101 & 44.5 & 7.87 & 
77.37 & 
93.56 &
78.32 (\textcolor{applegreen}{+0.95}) & 
94.21 (\textcolor{applegreen}{+0.65}) &
78.58 (\textcolor{applegreen}{+1.21}) &
94.42 (\textcolor{applegreen}{+0.86}) &
\textbf{79.59} (\textcolor{applegreen}{+2.22}) & 
\textbf{94.88} (\textcolor{applegreen}{+1.32}) \\[0.1em]

ResNet-152 & 60.2 & 11.61 & 
78.31 & 
94.06 &
79.24 (\textcolor{applegreen}{+0.92}) & 
94.72 (\textcolor{applegreen}{+0.66}) &
79.54 (\textcolor{applegreen}{+1.23}) &
94.98 (\textcolor{applegreen}{+0.92}) &
\textbf{80.74} (\textcolor{applegreen}{+2.43}) & 
\textbf{95.08} (\textcolor{applegreen}{+1.02}) \bstrut \\

\hline
DenseNet-121 & 8.0 & 2.90 & 
74.65 & 
92.17 &
75.88 (\textcolor{applegreen}{+1.23}) & 
92.92 (\textcolor{applegreen}{+0.75}) &
76.06 (\textcolor{applegreen}{+1.41}) &
93.16 (\textcolor{applegreen}{+0.99}) &
\textbf{77.29} (\textcolor{applegreen}{+2.64}) & 
\textbf{93.21} (\textcolor{applegreen}{+1.04}) \tstrut \\[0.1em]

DenseNet-161 & 28.7 & 7.85 & 
77.65 & 
93.80 &
78.72 (\textcolor{applegreen}{+0.93}) & 
94.41 (\textcolor{applegreen}{+0.61}) &
78.77 (\textcolor{applegreen}{+1.12}) &
94.53 (\textcolor{applegreen}{+0.73}) &
\textbf{80.10} (\textcolor{applegreen}{+2.35}) & 
\textbf{94.87} (\textcolor{applegreen}{+1.07}) \\[0.1em]

DenseNet-169 & 14.1 & 3.44 & 
76.00 & 
93.00 &
76.95 (\textcolor{applegreen}{+0.95}) & 
93.76 (\textcolor{applegreen}{+0.76}) &
77.19 (\textcolor{applegreen}{+1.19}) &
93.86 (\textcolor{applegreen}{+0.86}) &
\textbf{78.56} (\textcolor{applegreen}{+2.56}) & 
\textbf{94.23} (\textcolor{applegreen}{+1.23}) \bstrut \\
\hline

ResNeXt-50 32x4d & 25.0 & 4.29 & 
77.62 & 
93.70 &
78.48 (\textcolor{applegreen}{+0.86}) & 
93.37 (\textcolor{applegreen}{+0.67}) &
78.76 (\textcolor{applegreen}{+1.14}) &
94.48 (\textcolor{applegreen}{+0.78}) &
\textbf{79.98} (\textcolor{applegreen}{+2.36}) & 
\textbf{94.82} (\textcolor{applegreen}{+1.12}) \tstrut \\[0.1em]

ResNeXt-101 32x8d & 88.8 & 7.89 & 
79.31 & 
94.28 &
80.12 (\textcolor{applegreen}{+0.81}) & 
94.88 (\textcolor{applegreen}{+0.60}) &
80.57 (\textcolor{applegreen}{+1.26}) &
95.02 (\textcolor{applegreen}{+0.74}) &
\textbf{81.69} (\textcolor{applegreen}{+2.38}) & 
\textbf{95.51} (\textcolor{applegreen}{+1.23}) \bstrut \\

\hline
Wide-ResNet-50 & 68.9 & 11.46 & 
78.51 & 
94.09 &
79.52 (\textcolor{applegreen}{+1.01}) & 
94.85 (\textcolor{applegreen}{+0.76}) &
79.61 (\textcolor{applegreen}{+1.10}) &
94.92 (\textcolor{applegreen}{+0.83})  &
\textbf{80.24} (\textcolor{applegreen}{+1.73}) & 
\textbf{95.26} (\textcolor{applegreen}{+1.17})\tstrut \\[0.1em]

\end{tabular}
}
\end{table*}

\begin{table*}[t]
\caption{\textbf{Top-1 accuracy over runs} on ImageNet1K \cite{ILSVRC15} for original networks and those with \textit{e}mPool, \textit{e}DSCWPool and adaPool. We performed four runs for each combination of network and pooling type. The best run is denoted with (best). Best overall results in \textbf{bold}.}
\label{tab:ImageNet_multirun}
\centering
\resizebox{.95\textwidth}{!}{%
\begin{tabular}{l|ccc|c|ccc|c|ccc|c|ccc|c}
\hline
Pooling & \multicolumn{4}{c|}{Original (Baseline)} & \multicolumn{4}{c|}{\textit{e}MPool} & \multicolumn{4}{c|}{\textit{e}DSCWPool} & \multicolumn{4}{c}{adaPool} \tstrut \bstrut  \\
\cline{1-17}
\backslashbox{Model}{Run} & 1 & 2 & 3 & (best) & 1 & 2 & 3 & (best) &
1 & 2 & 3 & (best) & 1 & 2 & 3 & (best) \tstrut \bstrut \\
\hline
ResNet-18 & 
69.61 & 69.73 & 69.69 & 69.76 &
71.18 & 71.04 & 71.25 & 71.27 &
70.65 & 70.78 & 70.73 & 70.79 &
71.70 & 71.74 & 71.62 & \textbf{71.78} \tstrut \bstrut
\\
\hline
ResNet-34 & 
73.26 & 73.11 & 73.24 & 73.30 &
74.66 & 74.52 & 74.31 & 74.67 &
74.25 & 74.30 & 74.28 & 74.36 &
75.35 & 75.42 & 75.37 & \textbf{75.43}
\tstrut \bstrut
\\
\hline
ResNet-50 & 
76.01 & 75.97 & 76.04 & 76.15 &
77.26 & 77.24 & 77.19 & 77.35 &
77.35 & 77.26 & 77.23 & 77.38 &
78.36 & 78.38 & 78.41 & \textbf{78.42}
\tstrut \bstrut
\\
\end{tabular}
}
\end{table*}

\subsection{Image classification performance on ImageNet1K}
\label{sec:experiments::imagenet}

We test the assumption that a better preservation of information during downsampling with the exponential weighting method leads to an increase in image classification accuracy. Based on the results between average and max pooling and adaPool (Tables~\ref{tab:times_ssim_psnr} and \ref{tab:compute_flicker}), we replace the original pooling layers in ResNet \cite{he2016deep}, DenseNet \cite{huang2017densely}, ResNeXt \cite{xie2017aggregated} and wide-ResNet \cite{zagoruyko2016wide} networks with our exponential pooling method and test their performance on ImageNet1K. Results appear in Table~\ref{tab:ImageNet_scratch}. In Table~\ref{tab:ImageNet_multirun}, we summarize the results of four runs over different training seeds for three models to ensure fair comparisons. The highest accuracies are denoted by (best).

Overall, we notice that networks with their pooling layers replaced by adaPool yield improved accuracy rates. We provide a further discussion per CNN architecture.

\textbf{ResNet} \cite{he2016deep}. We report an average of 2.19\% top-1 and 1.33\% top-5 improvement on ResNet models when replacing their pooling layers with adaPool. Improvements in accuracy are also observed with replacements based on both \textit{e}MPool and \textit{e}DSCWPool with an average +1.17\% and +1.15\% top-1 accuracy, respectively. ResNet architectures include only a single pooling operation after the first convolution layer. The improvements from replacing only a single layer demonstrate the benefits of adaPool for image classification. In Table~\ref{tab:ImageNet_multirun}, we do not notice a significant divergence in accuracy over multiple runs on ResNet-18, ResNet-34, and ResNet-50 networks. On average, a replacement with adaPool can improve by +2.01\% the original ResNet-18 across runs, by +2.24\% on ResNet-34 and +2.38\% on ResNet-50.

\textbf{DenseNet} \cite{huang2017densely}. DenseNets include five pooling layers. Our replacements concern the maximum pooling layer after the first convolution and the four average pooling layers between dense blocks. The average top-1 accuracy gains based on layer replacements with adaPool are between 2.35--2.64\%. More modest increases are found for \textit{e}MPool and \textit{e}DSCWPool with +(0.93--1.23)\% and +(1.12--1.41)\%, respectively.

\textbf{ResNeXt} \cite{xie2017aggregated}. We achieve an average of 2.37\% top-1 and 1.17\% top-5 accuracy improvement with adaPool. An average gain of 1.20\% and 0.76\% for the top-1 and top-5 accuracies are observed with pooling layer replacement with \textit{e}DSCWPool. For \textit{e}MPool, these improvements are 0.83\% and 0.64\% for the top-1 and top-5 accuracies, respectively.

\textbf{Wide-Resnet-50} \cite{zagoruyko2016wide}. On Wide-ResNet-50, we observe the best top-1 accuracy of 80.24\% with a 1.73\% improvement when we replace the original pooling layers with adaPool. Gains in performance are also observed for \textit{e}MPool with +1.01\% and \textit{e}DSCWPool with +1.10\%.

\begin{table}[t]
\caption{\textbf{Pooling layer substitution} top-1 accuracy for a variety of pooling methods. Experiments were performed on ImageNet1K. Best results per network in \textbf{bold}.}
\label{tab:pooling_imagenet_tests}
\centering
\resizebox{.95\linewidth}{!}{%
\begin{tabular}{l|ccccccc}
\hline
\multirow{5}{*}{Pooling} & \multicolumn{7}{c}{Networks} \tstrut \\[0.2em]\cline{2-8}
\rule{0pt}{53pt} 
& \rotatebox{50}{\parbox{2.4mm}{ResNet$-$18}} 
& \rotatebox{50}{\parbox{2.4mm}{ResNet$-$34}} 
& \rotatebox{50}{\parbox{2.4mm}{ResNet$-$50}} 
& \rotatebox{50}{\parbox{2.4mm}{ResNeXt$-$50}}
& \rotatebox{50}{\parbox{2.4mm}{DenseNet$-$121}} 
& \rotatebox{50}{\parbox{2.4mm}{InceptionV1}}
& \bstrut \\\cline{1-7}
\multirow{2}{*}{Original (Baseline)} & (Max) & (Max) & (Max) & 
(Max) & (Avg+Max) & (Max) & \tstrut \\[0.1em]
& 69.76 & 73.30 & 76.15 & 
77.62 & 74.65 & 69.78 & \bstrut \\\cline{1-7}

Stochastic \cite{zeiler2013stochastic}  & 70.13 & 73.34 & 76.11 &
77.71 & 74.84 & 70.14 & \tstrut \\[0.1em]

S3 \cite{zhai2017s3pool} & 70.15 & 73.56 & 76.24 &
77.82 & 74.85 & 70.17 & \bstrut \\\cline{1-7}

$L_{p}$ \cite{lee2016generalizing} & 70.45 & 73.74 & 76.56 &
77.86 & 74.93 & 70.32 & \tstrut \\[0.1em]

Gate \cite{gulcehre2014learned} & 70.74 & 73.68 & 76.75 &
77.98 & 74.88 & 70.52 & \\[0.1em]

DPP \cite{saeedan2018detail} & 70.86 & 74.25 & 77.09 &
78.20 & 75.37 & 70.95 & \\[0.1em]

LIP \cite{gao2019lip} (drop-in) & 70.83 & 73.95 & 77.13 &
78.14 & 75.31 & 70.77 & \\[0.2em]

LIP \cite{gao2019lip} (multi) & 
71.42 & 74.86 & 78.19 &
79.25 & 76.64 & N/A & \bstrut \\\cline{1-7}

\textit{e}MPool \textbf{(ours)} & 71.27 & 74.67 & 77.35 & 78.48 & 75.88 & 71.43 & \tstrut \\[0.1em]

\textit{e}DSCWPool \textbf{(ours)} & 70.79 & 74.36 & 77.38 & 78.76 & 76.06 & 71.85 & \\[0.1em]

adaPool \textbf{(ours)} & \textbf{71.78} & \textbf{75.43} & \textbf{78.42} & \textbf{79.98} & \textbf{77.29} & \textbf{72.56} & \\[0.1em]

\end{tabular}
}
\end{table}

\begin{table*}[t]
\caption{\textbf{Object detection} bounding box AP results on MS COCO \texttt{test-dev} for models with original backbone networks and the same networks with pooling layers replaced by our exponential pooling layers. All models are pre-trained on ImageNet1K \cite{ILSVRC15}. Best results in \textbf{bold}.}
\begin{center}
    \centering
    \resizebox{\linewidth}{!}{%
    \begin{tabular}{l|c|ccc|ccc|ccc|ccc|ccc|ccc|ccc|ccc}
    \hline
    \multirow{2}{*}{Model} & \multirow{2}{*}{Backbone}  & \multicolumn{6}{c|}{Original (Baseline)} & \multicolumn{6}{c|}{\textit{e}MPool} &
    \multicolumn{6}{c|}{\textit{e}DSCWPool} & \multicolumn{6}{c}{adaPool} \tstrut\\
    & & AP & AP$_{50}\!$ & AP$_{75}\!$ & AP$_{S}\!$ & AP$_{M}\!$ & AP$_{L}\!$
      & AP & AP$_{50}\!$ & AP$_{75}\!$ & AP$_{S}\!$ & AP$_{M}\!$ & AP$_{L}\!$
      & AP & AP$_{50}\!$ & AP$_{75}\!$ & AP$_{S}\!$ & AP$_{M}\!$ & AP$_{L}\!$
      & AP & AP$_{50}\!$ & AP$_{75}\!$ & AP$_{S}\!$ & AP$_{M}\!$ & AP$_{L}\!$ \bstrut\\
    \hline
    \multirow{4}{*}{\rotatebox{90}{\parbox{2.1cm}{RetinaNet \cite{lin2017focal}}}} 
    & ResNet-18 & 28.3 & 48.7 & 31.6 & 
                 12.6 & 33.6 & 40.9 & 
                 
                 29.7 & 50.2 & 33.3 & 
                 14.1 & 35.2 & 42.6 & 
                 
                 28.9 & 49.6 & 32.8 &
                 13.8 & 34.7 & 41.5 &
                 
                 \textbf{31.2} & \textbf{51.4} & \textbf{34.7} &
                 \textbf{15.4} & \textbf{36.5} & \textbf{43.4} \tstrut \\[0.3em]
    
    & ResNet-34 & 31.6 & 50.8 & 33.9 & 
                 15.1 & 36.0 & 43.6 &
                 
                 32.8 & 52.1 & 35.5 & 
                 16.2 & 37.3 & 45.0 & 
                 
                 32.4 & 51.4 & 34.8 &
                 15.9 & 36.8 & 44.7 &
                 
                 \textbf{33.6} & \textbf{53.4} & \textbf{36.4} &
                 \textbf{16.9} & \textbf{38.2} & \textbf{44.7} \\[0.3em]
                                               
    & ResNet-50 & 34.0 & 52.5 & 36.5 & 
                 17.0 & 37.4 & 45.1 &
                 
                 34.9 & 53.4 & 37.6 & 
                 18.0 & 38.5 & 46.4 & 
                 
                 34.6 & 53.1 & 37.2 &
                 17.7 & 38.2 & 46.1 &
                 
                 \textbf{35.6} & \textbf{53.9} & \textbf{38.0} &
                 \textbf{18.4} & \textbf{39.1} & \textbf{47.2} \\[0.3em]
                 
    & ResNet-101 & 39.1 & 59.1 & 42.3 & 
                  21.8 & 42.7 & 50.2 &
                  
                  39.8 & 59.9 & 43.3 & 
                  22.4 & 43.5 & 51.1 & 
                  
                  40.1 & 60.3 & 43.7 &
                  22.6 & 43.9 & 51.4 &
                  
                  \textbf{40.8} & \textbf{61.6} & \textbf{44.8} &
                  \textbf{23.7} & \textbf{44.8} & \textbf{52.5} \bstrut \\
    \hline
    \multirow{3}{*}{\rotatebox{90}{\parbox{1.2cm}{Mask R-CNN \cite{he2017mask}}}} 
    
    & ResNet-34 & 32.9 & 53.6 & 32.7 & 
                 14.5 & 35.1 & 43.2 &
                 34.0 & 54.8 & 34.1 & 
                 15.7 & 36.6 & 44.6 & 
                 33.8 & 54.1 & 33.6 &
                 15.3 & 36.2 & 44.0 &
                 \textbf{35.7} & \textbf{56.9} & \textbf{36.4} &
                 \textbf{16.8} & \textbf{38.6} & \textbf{46.5} \tstrut \\[0.2em]
               
    & ResNet-50 & 33.6 & 55.2 & 35.3 & 
                 15.4 & 36.8 & 45.3 &
                 34.5 & 56.2 & 36.4 & 
                 16.2 & 37.7 & 46.3 & 
                 34.4 & 56.2 & 36.3 &
                 16.3 & 37.5 & 46.2 &
                 \textbf{36.3} & \textbf{57.5} & \textbf{36.9} &
                 \textbf{17.1} & \textbf{39.0} & \textbf{47.3} \\[0.2em]
    
    & ResNet-101 & 38.2 & 60.3 & 41.7 & 
                  20.1 & 41.1 & 50.2 &
                  39.0 & 61.1 & 42.6 & 
                  20.9 & 42.0 & 51.3 & 
                  39.5 & 61.7 & 43.1 &
                  21.5 & 42.8 & 51.9 &
                  \textbf{42.4} & \textbf{62.8} & \textbf{45.1} &
                  \textbf{24.5} & \textbf{45.6} & \textbf{52.8} \\[0.2em]
    
    \end{tabular}
    }
\end{center}%
\label{tab:segmentation_results}
\end{table*}

\subsection{Comparison with alternative pooling methods}
\label{sec:experiments::pooling_methods}

We provide quantitative comparisons between different pooling methods over six different models in Table~\ref{tab:pooling_imagenet_tests}. We systematically replaced the pooling layers of the original model (baseline). For LIP, we consider both drop-in replacements, in line with the rest of our experiments, as well as multiple replacements following the LIP-ResNet and LIP-DenseNet architectures of the paper~\cite{gao2019lip}. Non-adaptive \textit{e}MPool and \textit{e}DSCWPool still outperform stochastic methods while the obtained accuracies are similar to those of learnable methods. Across the tested architectures, adaPool outperforms other learnable and stochastic pooling methods. The largest overall margins are observed for InceptionV1 with improvements over other methods in the range of 1.61--2.78\% and on DenseNet-121 (0.65--2.64\%).

\subsection{Object detection performance on MS COCO}
\label{sec:experiments::detection}

To investigate the merits of our proposed exponentially-weighted pooling on encapsulating relevant local information, we present results for object detection on MS COCO \cite{lin2014microsoft} in Table~\ref{tab:segmentation_results}. We use RetinaNet \cite{lin2017focal} and Mask-RCNN \cite{he2017mask} with several different backbone networks. We chose these two models based on their wide popularity. Overall, we observe that both \textit{e}MPool and \textit{e}DSCWPool come with average precision (AP) improvements of 1.00\% and 0.86\%, respectively. A 2.40\% increase over the original models is observed for adaPool. Similar trends in AP are also visible for AP$_{50}$ and AP$_{75}$, demonstrating that adaPool does not only benefit tasks that rely primarily on general features such as classification, but also provides a performance boost for local-based feature tasks such as object detection.

\begin{table}[t]
\caption{\textbf{Action recognition top-1 and top-5 accuracies for HACS, K-700 and UCF-101}. Models are trained on HACS and fine-tuned on K-700 and UCF-101, except for ir-CSN-101 and SF r3d-50 (see text). N/A means no trained model was provided. Best results in \textbf{bold}.
}
\label{tab:video_accuracies}
\begin{threeparttable}[t]
\centering
\resizebox{\linewidth}{!}{%
\begin{tabular}{l|c|cc|cc|cc}
\hline
\multicolumn{1}{c|}{\multirow{2}{*}{Model}} & 
\multicolumn{1}{c|}{FLOPs} &
\multicolumn{2}{c|}{HACS} &
\multicolumn{2}{c|}{K-700} &
\multicolumn{2}{c}{UCF-101} \tstrut \\
&
(G) &
top-1 & top-5 & 
top-1 & top-5 &
top-1 & top-5 \bstrut \\
\hline
r3d-101 \cite{kataoka2020would}$^{**}$ &
78.5 &
80.49 & 95.18 &
52.58 & 74.63 & 
95.76 & 98.42 \tstrut \\
r(2+1)d-50 \cite{tran2018closer}$^{**}$ &
50.0 &
81.34 & 94.51 &
49.93 & 73.40 & 
93.92 & 97.84 \\
I3D \cite{carreira2017quo}$^{\ddagger *}$ &
55.3 &
79.95 & 94.48 &
53.01 & 69.19 & 
92.45 & 97.62 \\
ir-CSN-101 \cite{tran2019video}$^{\ddagger \dagger}$ &
17.3 &
N/A & N/A &
54.66 & 73.78 &
95.13 & 97.85 \\
SRTG \cite{stergiou2021learn}$^{\dagger \dagger}$ &
78.7 &
81.66 & 96.37 &
56.46 & 76.82 &
97.32 & 99.56 \\
SF r3d-50 \cite{feichtenhofer2019slowfast}$^{\ddagger \dagger}$ &
36.7 &
N/A & N/A &
56.17 & 75.57 &
94.62 & 98.75 \\

MTNet$_L$ \cite{stergiou2021multi}$^{\dagger \dagger}$ &
17.6 &
86.62 & 96.68 &
63.31 & 84.14 &
97.38 & 99.23 \bstrut \\

\hline
\cite{tran2018closer} w/ adaPool &
53.2 &
81.13 & 94.96 &
50.87 & 74.06 &
94.21 & 97.76 \tstrut \\
\cite{stergiou2021learn} w/ adaPool &
78.7 &
84.37 & 97.84 &
58.62 & 78.56 &
98.53 & \textbf{99.86} \\
\cite{stergiou2021multi} w/ adaPool &
17.8 &
\textbf{87.83} & \textbf{98.21} & 
\textbf{64.67} & \textbf{84.78} &
\textbf{98.60} & 99.74 \\
\end{tabular}%
}
 \begin{tablenotes}
    \item[$**$] re-implemented models trained from scratch.  
    \item[$\dagger \dagger$] models and weights from official repositories. \item[$\ddagger *$] unofficial models trained from scratch. 
    \item[$\ddagger \dagger$] models from unofficial repositories with official weights.
   \end{tablenotes}
\end{threeparttable}%
\end{table}

\subsection{Video classification performance}
\label{sec:experiments::video}

We evaluate our pooling operators on spatio-temporal data by focusing on the task of action recognition in videos. The accurate classification and representation of space-time features stands as a major challenge in the field of video understanding \cite{stergiou2019analyzing}.

The majority of space-time networks are based on the extension of 2D convolutions to 3D to include the temporal dimension. Stacks of frames are used as inputs. Similarly, the only modification in our method is the inclusion of the temporal dimension in kernel region \textbf{R}.

For our tests, we first train models from scratch on HACS \cite{zhao2019hacs} using the author implementations. These models are then used to initialize the weights for the Kinetics-700 and UCF-101 tests. SlowFast (SF) \cite{feichtenhofer2019slowfast} and ir-CSN-101 \cite{tran2019video} are the only two models that use different initialization weights, with ir-CSN-101 pre-trained on IG65M and SF on ImageNet.

We report in Table~\ref{tab:video_accuracies} the performance of three spatio-temporal CNNs with pooling layers replaced by adaPool. We observe state-of-the-art performance using MTNet$_{L}$ with adaPool on HACS and Kinetics-700, with 87.83\% and 64.67\% top-1 accuracies, respectively. This corresponds to an increase of 1.21\% and 1.36\% over the same networks with the original pooling layers. This also comes with negligible additional GFLOPs (+0.2). On UCF-101, we show that both MTNet$_{L}$ and SRTG r3d-101 with adaPool outperform the original and other top-performing models. Increases in top-1 performance are also observed for SRTG r3d-101 with +2.71\% on HACS and +1.47\% on Kinetics-700.

These results further demonstrate that the simple replacement of a pooling operator by adaPool consistently results in a modest but important performance gain. Even for the almost saturated performance on UCF-101, using adaPool results in a performance increase of 1.22\% on MTNet$_{L}$.

\begin{table}[t]
\caption{\textbf{Image super-resolution with $\mathbf{\times 2}$ and $\mathbf{\times 4}$ upsampling}. Best and second best results in \textbf{bold} and \underline{underlined}.}
\label{tab:image_superres}
\centering
\resizebox{\linewidth}{!}{%
\setlength\tabcolsep{1.8pt}
\begin{tabular}{l|l|ccc|cc|ccc}
\hline
\multicolumn{1}{c|}{\multirow{2}{*}{Scale}} &
\multicolumn{1}{c|}{\multirow{2}{*}{Model}} &
\multicolumn{3}{c|}{Urban100 \cite{xue2019video}} &
\multicolumn{2}{c|}{Manga109 \cite{matsui2017sketch}} &
\multicolumn{3}{c}{B100 \cite{martin2001database}} \tstrut\\
& & PSNR & SSIM & LPIPS & PSNR & SSIM & PSNR & SSIM & LPIPS \bstrut  \\
\hline
\multirow{7}{*}{2x} &
Bicubic & 26.88 & 0.8431 & 0.383$^{*}$ & 30.80 & 0.9339 & 29.56 & 0.8316 & 0.396$^{*}$ \tstrut \\[.2em]
& SRCNN \cite{dong2014learning} & 29.50 & 0.8946 & N/A & 35.60 & 0.9663 & 31.36 & 0.8879 &  N/A \\[.2em]
& RCAN \cite{zhang2018image} & 33.34 & 0.9384 & 0.046$^{*}$ & 39.44 & 0.9786 & 32.41 & 0.9027 & 0.064$^{*}$ \tstrut \\[.2em]
& SAN \cite{dai2019second} & 33.10 & 0.9370 & N/A & 39.32 & 0.9792 & 32.42 & 0.9028 & N/A \\[.2em]
& HAN+ \cite{niu2020single} & 33.53 & 0.9398 & 0.038$^{*}$ & 39.62 & 0.9787 & 32.41 & 0.9027 & 0.060$^{*}$ \bstrut \\[.2em]
\cline{2-10}
& RCAN w/ adaP/U & \underline{33.58} & \underline{0.9456} & \underline{0.036} & \underline{39.67} & \underline{0.9834} & \underline{32.63} & \underline{0.9103} & \underline{0.057} \tstrut \\[.2em]
& HAN+ w/ adaP/U & \textbf{33.72} & \textbf{0.9469} & \textbf{0.027} & \textbf{39.82} & \textbf{0.9841} & \textbf{32.79} & \textbf{0.9187} & \textbf{0.051} \bstrut \\
\hline
\multirow{8}{*}{4x} &
Bicubic & 23.14 & 0.6577 & 0.473 & 24.89 & 0.7866 & 25.96 & 0.6675 & 0.525 \tstrut \\[.2em]
& SRCNN \cite{dong2014learning} & 24.52 & 0.7221 & N/A & 27.58 & 0.8555 & 26.90 & 0.7101 & N/A \\[.2em]
& RCAN \cite{zhang2018image} & 26.82 & 0.8087 & 0.098$^{*}$ & 31.22 & 0.9173 & 27.77 & 0.7436 & 0.121$^{*}$ \\[.2em]
& SFTGAN \cite{wang2018recovering} & 25.51 & 0.7549 & 0.177 & N/A & N/A & 27.13 & 0.7354 & 0.178 \\[.2em]
& SAN \cite{dai2019second} & 26.79 & 0.8068 & N/A & 31.18 & 0.9169 & 27.78 & 0.7436 & N/A \\[.2em]
& SRGAN \cite{ledig2017photo} & 25.50 & 0.7485 & 0.198 & N/A & N/A & 27.09 & 0.7360 & 0.171 \\[.2em]
& HAN+ \cite{niu2020single} & 27.02 & 0.8131 & 0.093$^{*}$ & 31.73 & 0.9207 & 27.85 & 0.7454 & 0.105$^{*}$ \bstrut \\
\cline{2-10}
& RCAN w/ adaP/U & \underline{27.24} & \underline{0.8195} & \underline{0.089} & \underline{31.78} & \underline{0.9243} & \textbf{28.11} & \underline{0.7482} & \textbf{0.093} \tstrut \\[.2em]
& HAN+ w/ adaP/U & \textbf{27.96} & \textbf{0.8246} & \textbf{0.066} & \textbf{32.30} & \textbf{0.9286} & \underline{28.06} & \textbf{0.7513} & \underline{0.095} \\[.2em]
\end{tabular}
}
\end{table}

\begin{table}[t]
\caption{\textbf{Qualitative frame interpolation results on Vimeo90K, Middlebury and Inter4K}. N/A indicates that the results were not provided in the original works. Best results in \textbf{bold}.}
\label{tab:frame_interpolation_results}
\centering
\resizebox{\linewidth}{!}{%
\begin{tabular}{l|ccc|ccc|ccc}
\hline
\multicolumn{1}{c|}{\multirow{2}{*}{Model}} &
\multicolumn{3}{c|}{Vimeo90K \cite{xue2019video}} &
\multicolumn{3}{c|}{Middlebury \cite{baker2011database}} &
\multicolumn{3}{c}{Inter4K (4K, $30 \! \rightarrow \! 60$fps)} \tstrut\\[0.2em]
 & PSNR & SSIM & LPIPS & PSNR & SSIM & LPIPS & PSNR & SSIM & LPIPS \bstrut \\
\hline
DAIN \cite{bao2019depth} 
& 34.70 & 0.964 & 0.022 
& 36.70 & 0.965 & 0.017 
& 35.48 & 0.959 & 0.021 \tstrut \\[.2em]
CAIN \cite{choi2020channel} & 34.65 & 0.959 & 0.020 & 35.11 & 0.951 & 0.019 & 34.92 & 0.953 & 0.019 \\[.2em]
BMBC \cite{park2020bmbc} & 35.06 & 0.964 & 0.015 & 36.79 & 0.965 & 0.015 & 35.76 & 0.966 & 0.015 \\[.2em]
XVFI \cite{sim2021xvfi} & 34.27 & 0.971 & N/A & N/A & N/A & N/A & 35.28 & 0.969 & 0.018 \\[.2em]
CDFI \cite{ding2021cdfi} & 35.17 & 0.964 & 0.010 & 37.14 & 0.966 & 0.007  & 36.31 & 0.967 & 0.010 \bstrut \\
\hline
\cite{bao2019depth} w/ adaP/U & 34.96 & 0.968 & 0.017 & 36.82 & 0.968 & 0.015 & 35.73 & 0.964 & 0.012 \tstrut \\[.2em]
\cite{ding2021cdfi} w/ adaP/U & \textbf{35.23} & \textbf{0.972} & \textbf{0.008} & \textbf{37.22} & \textbf{0.970} & \textbf{0.006}  & \textbf{36.57} & \textbf{0.972} & \textbf{0.007} \\[.2em]
\end{tabular}
}
\end{table}

\begin{table}[t]
\caption{\textbf{Frame interpolation and super-resolution with CDFI on Inter4K}. The resolutions and fps of the original and processed videos are indicated in the second column. Best results in \textbf{bold}.}
\label{tab:frame_interpolation_superres_results}
\centering
\resizebox{.9\linewidth}{!}{%
\begin{tabular}{l|c|ccc}
\hline
\multicolumn{1}{c|}{\multirow{2}{*}{Scale}} &
\multicolumn{1}{c|}{\multirow{2}{*}{Resolution and fps conversions}} &
\multicolumn{3}{c}{Measures} \tstrut \\
& & PSNR & SSIM & LPIPS \bstrut \\
\hline
\multirow{4}{*}{2x}
& nHD\@15fps $\rightarrow$ HD\@30fps & 33.95 & 0.936 & 0.018 \tstrut \\[.2em]
& qHD\@24fps $\rightarrow$ FHD\@50fps & 33.91 & 0.928 & 0.020 \\[.2em]
& HD\@30fps $\rightarrow$ QHD\@60fps & 33.87 & 0.925 & 0.021 \\[.2em]
& FHD\@30fps $\rightarrow$ UHD\@60fps & 33.81 & 0.918 & 0.021 \bstrut \\
\hline
\multirow{2}{*}{4x} 
& nHD\@15fps $\rightarrow$ QHD\@60fps & 25.32 & 0.822 & 0.028 \tstrut \\[.2em]
& qHD\@15fps $\rightarrow$ UHD\@60fps & 25.38 & 0.819 & 0.031 \\[.2em]

\end{tabular}
}
\end{table}

\subsection{Image super-resolution and frame interpolation results}
\label{sec:experiments::superres_iner}

In order to assess the benefits of re-using the learned adaPool weights in adaUnPool, we experiment on image super-resolution, video frame interpolation, and their combination. For each task we replace pooling layers with adaPool and the respective bilinear interpolation with adaUnPool.

Our comparisons on image super-resolution are shown in Table~\ref{tab:image_superres}. Both RCAN \cite{zhang2018image} and HAN+ \cite{niu2020single} perform favorably with down and up-sampling layers substituted by ada(Un)Pool. We observe that, in both cases of $2 \times$ and $4 \times$ image upsampling, our converted networks not only outperform their original implementations, but also other methods.

We demonstrate the merits of replacing all pooling and interpolation layers with ada(Un)Pool for frame interpolation in Table~\ref{tab:frame_interpolation_results}. The two converted networks, DAIN \cite{bao2019depth} and CDFI \cite{ding2021cdfi}, produce improved results across the tested datasets. CDFI with adaPool and adaUnPool yields state-of-the-art results on both Vimeo90K and Middlebury as well as on our Inter4K for 4K video interpolation from 30 to 60 fps.

We also perform benchmarking tests on Inter4K with CDFI+ada(Un)Pool for the combined task of frame super-resolution and interpolation. Our findings are reported in Table~\ref{tab:frame_interpolation_superres_results}. Overall, we observe only slight degradation in performance on high-resolution, high-frame-rate conversions.

\section{Ablation studies} 
\label{sec:ablation}

In this section, we investigate the impact of different design choices for adaPool. We initially consider the effect of setting the $\boldsymbol \beta$ weight mask as trainable parameter or as constant value (Section~\ref{sec:ablations::adapool_param}). Additionally, we provide results on pooling layer replacements on the InceptionV3 \cite{szegedy2016rethinking} (Section~\ref{sec:ablations::multi-layer}), evaluate the performance over fusion and pooling method substitutions (Section~\ref{sec:ablations::fusion_pooling}), and compare against attention-based methods converted to downsampling (Section~\ref{sec:ablations::attention}). Finally, we present qualitative visualizations of network saliency and the feature embedding space over original and adaPool-replaced models (Section~\ref{sec:ablations::visualizations}). Unless otherwise specified, experiment settings follow those described in Section~\ref{sec:experiments::settings}.

\begin{table}[t]
\caption{\textbf{Effect of $\boldsymbol{\beta}$ on ImageNet1K image classification}. Larger values of $\boldsymbol{\beta}$ correspond to stronger reliance on \textit{e}DSCWPool while smaller $\boldsymbol{\beta}$ values prioritize \textit{e}MPool. Best results are in \textbf{bold} while second best results are \underline{underlined}.}
\label{tab:beta_replacements}
\begin{center}
    \centering
    \resizebox{.9\linewidth}{!}{%
    \begin{tabular}{l|c|cc|cc|cc}
    \hline
    \multirow{2}{*}{Mode} & \multirow{2}{*}{$\boldsymbol{\beta}$ value} & \multicolumn{2}{c|}{ResNet-18} & \multicolumn{2}{c|}{ResNet-34} & \multicolumn{2}{c}{InceptionV3} \tstrut\\[0.1em] 
    & & top-1 & top-5 & top-1 & top-5 & top-1 & top-5 \bstrut\\
    \hline
    \multirow{7}{*}{\rotatebox{90}{Constant}} 
    & $\boldsymbol{\beta}$=1/8 & 71.31 & 90.21 & \underline{74.83} & \underline{92.42} & 78.98 & 93.77 \tstrut \\[0.1em]
    & $\boldsymbol{\beta}$=1/4 & \underline{71.34} &\underline{90.26} & 74.76 & 92.38 & 79.23 & 93.84 \\[0.1em]
    & $\boldsymbol{\beta}$=3/8 & 71.31 & 90.19 & 74.63 & 92.34 & 79.35 & 93.92 \\[0.1em]
    & $\boldsymbol{\beta}$=1/2 & 71.28 & 90.07 & 74.56 & 92.31 & 79.54 & 93.89 \\[0.1em]
    & $\boldsymbol{\beta}$=5/8 & 71.16 & 90.02 & 74.48 & 92.28 & 79.68 & 94.01 \\[0.1em]
    & $\boldsymbol{\beta}$=3/4 & 71.19 & 89.95 & 74.38 & 92.16 & 79.97 & 94.05 \\[0.1em]
    & $\boldsymbol{\beta}$=7/8 & 71.04 & 89.96 & 74.41 & 92.20 & \underline{80.16} & \underline{94.19} \bstrut \\
    \hline
    \multicolumn{2}{c|}{trainable}& \textbf{71.78} & \textbf{90.65} & \textbf{75.43} & \textbf{92.87} & \textbf{81.34} & \textbf{94.57} \tstrut \\
    
    \end{tabular}
    }
\end{center}%
\end{table}

\subsection{Effect of $\beta$ weight mask}
\label{sec:ablations::adapool_param}

In order to study how different combinations of the approximated maximum and average effect our proposed adaPool method, we present results in Table~\ref{tab:beta_replacements} on ImageNet1K with several constant $\boldsymbol{\beta}$ values and study the performance gains when $\boldsymbol{\beta}$ is converted to a trainable weight mask. 

Overall, the trainable setting provides the best performance across all three tested networks. The performance improvement of the trainable weight mask over the best-performing constant value becomes more apparent in complex architectures. In ResNet-18 the difference in top-1 is 0.44\% while in InceptionV3 it becomes 1.18\%. We provide further parameterization-based ablations in Appendix~\ref{sec:ablations::beta_param}.

\begin{table}[t]
\caption{\textbf{Progressive layer substitution for InceptionV3 on ImageNet1K}. Column numbers refer to the number of replaced pooling layers, marked with \checkmark. Best results in \textbf{bold}.}
\label{tab:inception_tests_progressive}
\centering
\resizebox{\linewidth}{!}{%
\begin{tabular}{lcccccccc}
\hline
\multirow{2}{*}{Layer} & \multicolumn{8}{c}{Pooling layer substitution with adaPool} \tstrut \bstrut \\\cline{2-9}
& N & I & II & III & IV & V & VI & VII \tstrut \bstrut\\
\hline
$pool_{1}$ & & \checkmark & \checkmark & \checkmark & \checkmark & \checkmark & \checkmark & \checkmark \tstrut \\ 
$pool_{2}$ & & & \checkmark & \checkmark & \checkmark & \checkmark & \checkmark & \checkmark \\
$mixed \; 5_{b-d}$ & & & & \checkmark & \checkmark & \checkmark & \checkmark & \checkmark \\
$mixed \; 6_{a}$ & & & & & \checkmark & \checkmark & \checkmark & \checkmark \\
$mixed \; 6_{b-e}$ & & & & & & \checkmark & \checkmark & \checkmark \\
$mixed \; 7_{a}$ & & & & & & & \checkmark & \checkmark \\
$mixed \; 7_{b-d}$ & & & & & & & & \checkmark \bstrut \\
\hline
Top-1 (\%) & 77.45 & 78.34 & 78.89 & 79.32 & 79.78 & 80.21 & 80.54 & \textbf{81.34} \tstrut \\
Top-5 (\%) & 93.56 & 93.77 & 93.92 & 94.05 & 94.17 & 94.26 & 94.32 & \textbf{94.57} \\
\end{tabular}
}
\end{table}

\subsection{Layer-wise ablation on InceptionV3}
\label{sec:ablations::multi-layer}

To understand the effect of adaPool at different network depths, we hierarchically ablate over pooling layers of the InceptionV3 architecture. This choice is primarily based on the Inception block's structure that includes pooling operations. This allows for a per-block evaluation of the change in the pooling operator.

From results summarized in Table~\ref{tab:inception_tests_progressive}, we observe that we can expect an average increase of 0.56\% in top-1 accuracy with each additional replacement of an original pooling operation by adaPool. While the performance gains are systematic, the largest improvements are observed for replacements over the first pooling operation after the initial convolutional layer ($pool_{1}$) with a 0.89\% jump in accuracy, and at the final Inception block ($mixed7_{b-d}$) with a 0.80\% increase. We thus demonstrate that adaPool yields accuracy improvement through its adaptive weighting, regardless of the network depth and number of channels.

\begin{table}[t]
\caption{\textbf{Top-1 accuracy over runs on ImageNet1K based on different pooling and pooling combination methods}. A ResNet-18 is used for all experiments. Top results are in \textbf{bold} and the best result per fusion method is \underline{underlined}.}
\label{tab:ImageNet_ada_combi}
\centering
\resizebox{\textwidth}{!}{%
\begin{tabular}{l|ccc|c}
\hline
\backslashbox{Fusion}{Pooling} & avg+max & avg+\textit{e}M & \textit{e}DSCW+max & \textit{e}DSC+\textit{e}M \tstrut \bstrut \\
\hline
mixed & 
70.37 & 70.73 &
70.65 & \underline{71.08} \tstrut \\[.1em]
gate & 
71.04 & 71.25 &
71.32 & \underline{71.44} \bstrut \\
\hline
adaptive \textbf{(ours)} & 
71.42 & 71.56 &
71.53 & \textbf{71.78} \tstrut \\[.1em]
\end{tabular}
}
\end{table}

\subsection{Pooling combinations over fusion methods}
\label{sec:ablations::fusion_pooling}

We provide comparisons over additional pooling methods and fusion strategies proposed in \cite{lee2016generalizing}. The \textit{mixed} pooling fusion strategy corresponds to using a single parameter to fuse the pooling methods used. This can be considered as a special case of adaptive pooling in which $|\boldsymbol \beta| = 1$. The \textit{gate} fusion method uses a learned parameter to select either of the two used pooling methods. In addition to our \textit{e}DSCW+\textit{e}M combination, we also test fusion strategies with average/maximum pooling.

Our comparisons are shown in Table~\ref{tab:ImageNet_ada_combi}. The combination of the smooth approximated average and maximum performs favorably over the different average or maximum-based combinations. We also observe that the use of a parameter mask through adaptive fusion helps to improve performance.

\begin{table}[t]
\resizebox{\textwidth}{!}{
\caption{\textbf{Comparison of adaPool to attention-based downsampling for DenseNet-121 on ImageNet1K}, with SE \cite{hu2018squeeze}, CBAM \cite{woo2018cbam}, and MSA \cite{li2022mvitv2}. Best results are in \textbf{bold}.}
\label{tab:attn_ablate}
\centering
\begin{tabular}{l|llcc}\hline
\multicolumn{1}{c|}{Method} & top-1 & top-5 & +Params & +FLOPs \tstrut \bstrut \\ \hline
\multicolumn{5}{l}{\textit{Fixed approaches}} \\
AvgPool (Baseline) & 74.65 & 92.17 & - & - \\
\textit{e}M/SoftPool \cite{stergiou2021refining} & 75.88 & 92.92 & - & - \\
\textit{e}DSCWPool & 76.06 & 93.16 & - & - \bstrut \\
\hline
\multicolumn{5}{l}{\textit{Learned approaches}} \tstrut \\
AvgPool + SE \cite{hu2018squeeze} & 76.32 & 93.06 & +43.9K & +0.2G \\
\textit{e}M/SoftPool + SE & 76.45 & 93.09 & +43.9K & +0.2G \\
AvgPool + CBAM \cite{woo2018cbam} & 77.03 & 93.16 & +44.3K & +0.5G \\
\textit{e}M/SoftPool + CBAM & 77.11 & 93.18 & +44.3K & +0.5G \\
AvgPool + MSA \cite{li2022mvitv2} & 77.38 & 93.27 & +1.4M & +2.5G \\
\textit{e}M/SoftPool + MSA & \textbf{77.51} & \textbf{93.36} & +1.4M & +2.5G \\
adaPool \textbf{(ours)} & 77.29 & 93.21 & \textbf{+4.2K} & \textbf{+1.5M} \\
\end{tabular}
}
\end{table}

\subsection{Comparisons to attention-based downsampling}
\label{sec:ablations::attention}

The recent introduction of attention-based methods has shown great promise for a range of high-level vision tasks. We therefore also investigate the usability of three different attention-based approaches by adapting them for downsampling. We test the channel-wise Squeeze-and-Excitation (SE) \cite{hu2018squeeze} attention module, the locally-applied Convolutional Block Attention Module (CBAM) \cite{woo2018cbam}, and the Multiscale Self Attention module (MSA) \cite{li2022mvitv2} that uses global attention over spatially reduced $\textbf{KQV}$ linear projections of the input. The tested modules are converted for downsampling by pooling after (SE, CBAM) or before (MSA) the attention modules.

From the results presented in Table~\ref{tab:attn_ablate}, we observe that our proposed adaPool is substantially more efficient than any attention-based method with only requiring +1.5 additional MFLOPs and 4.2K parameters. AdaPool shows to perform favorably compared to SE-based and CBAM-based pooling methods while a small decrease in performance is observed in comparison to MSA with average or SoftPool. We note that the performance-to-computational complexity trade-off between adaPool and MSA-based pooling is substantial, with MSA requiring 1,600 more FLOPs than adaPool. For DenseNet-121 the computational burden with using MSA-based pooling is $~30\%$ of the total number of FLOPs used by the model.

\begin{figure*}[!htb]
    \RawFloats
    \begin{minipage}[t]{.323\linewidth}
    \centering
        \includegraphics[trim=65px 40px 50px 45px, clip,width=\linewidth]{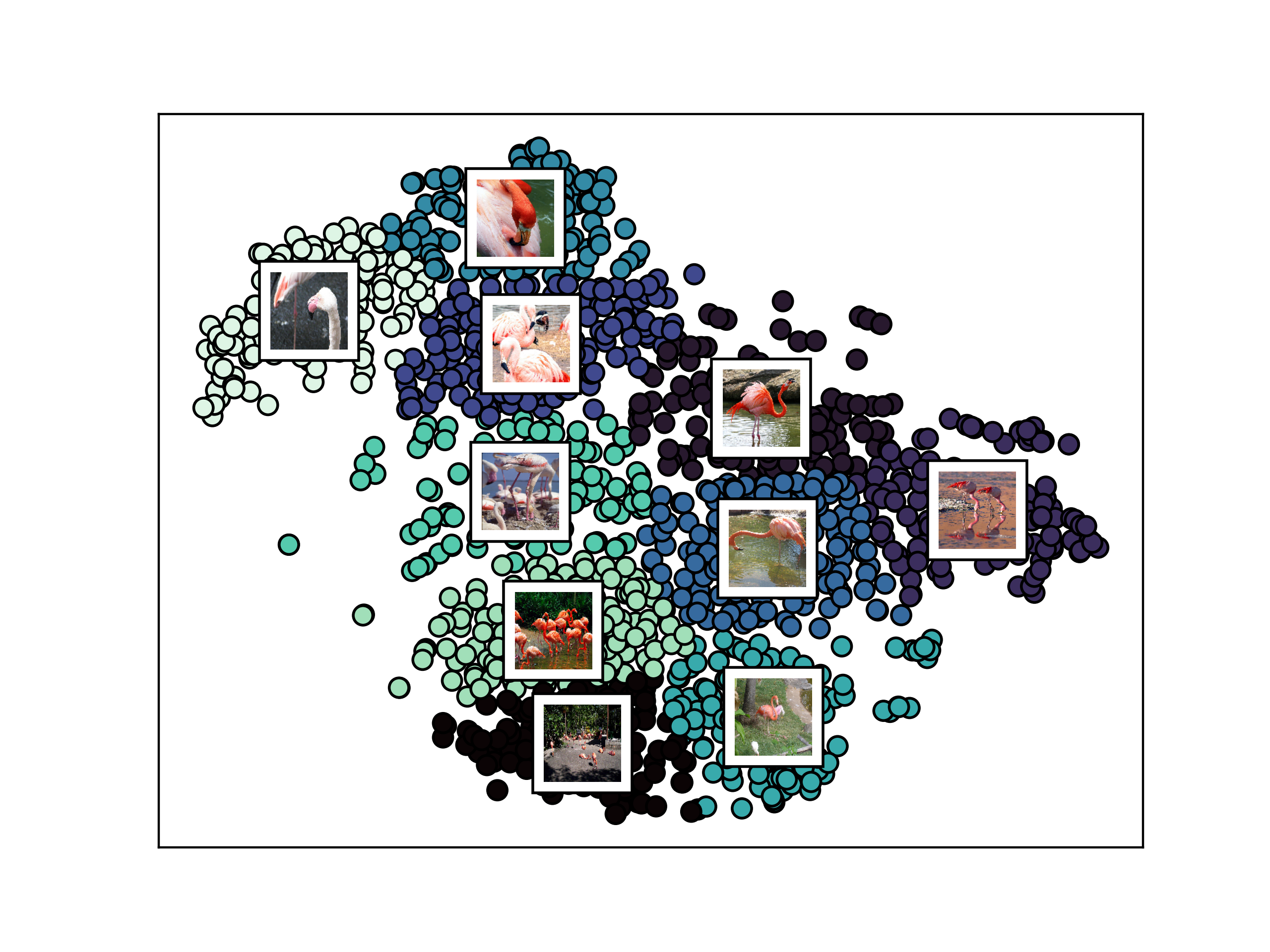}
        
        \includegraphics[trim=65px 40px 50px 45px, clip,width=\linewidth]{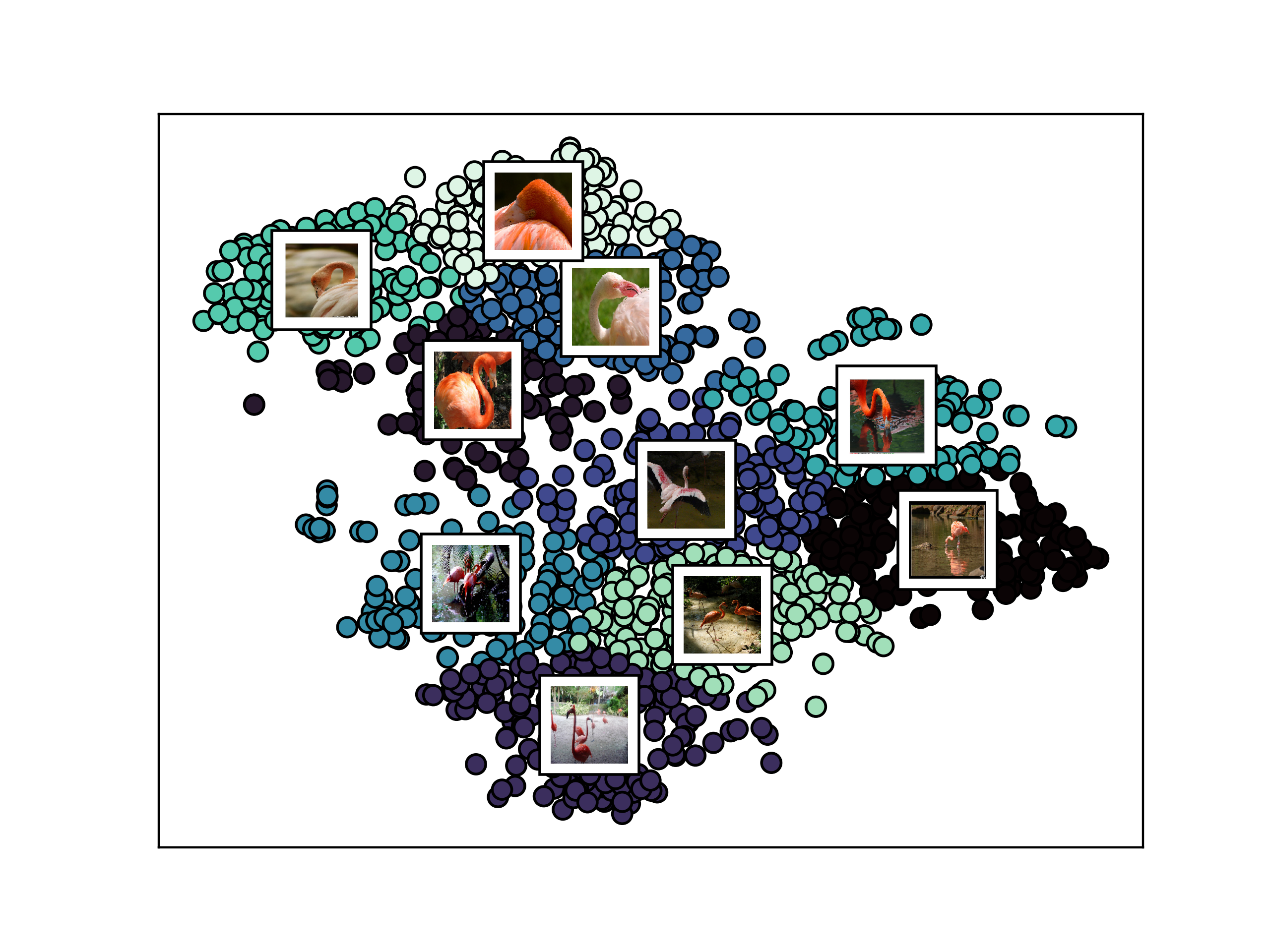}
        \caption*{a. Class ``flamingo''}
        \label{fig:tsne_original_a}
    \end{minipage}
    \hfill
    \begin{minipage}[t]{.323\linewidth}
    \centering
        \includegraphics[trim=65px 40px 50px 45px, clip,width=\linewidth]{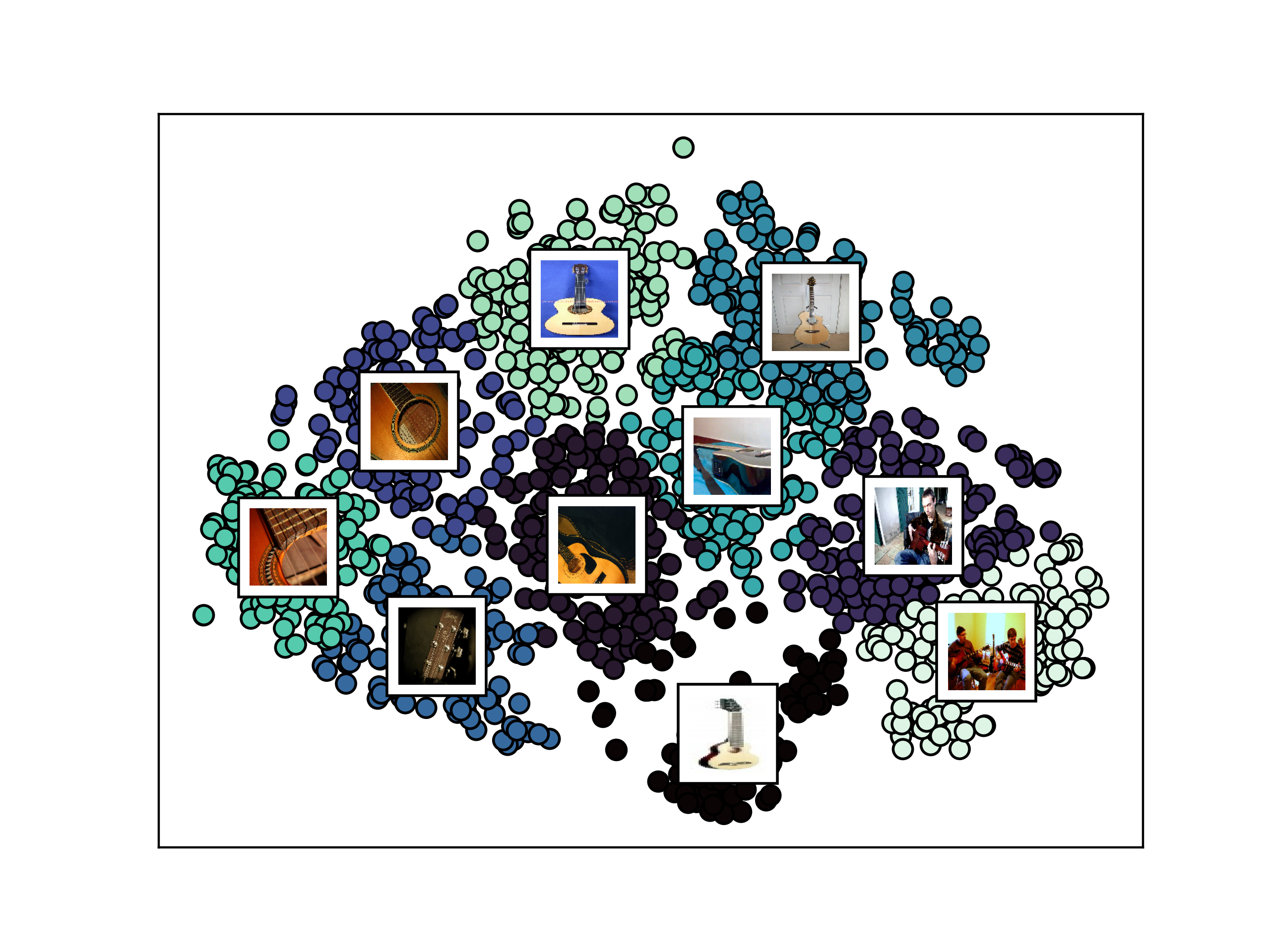}
        
        \includegraphics[trim=65px 40px 50px 45px, clip,width=\linewidth]{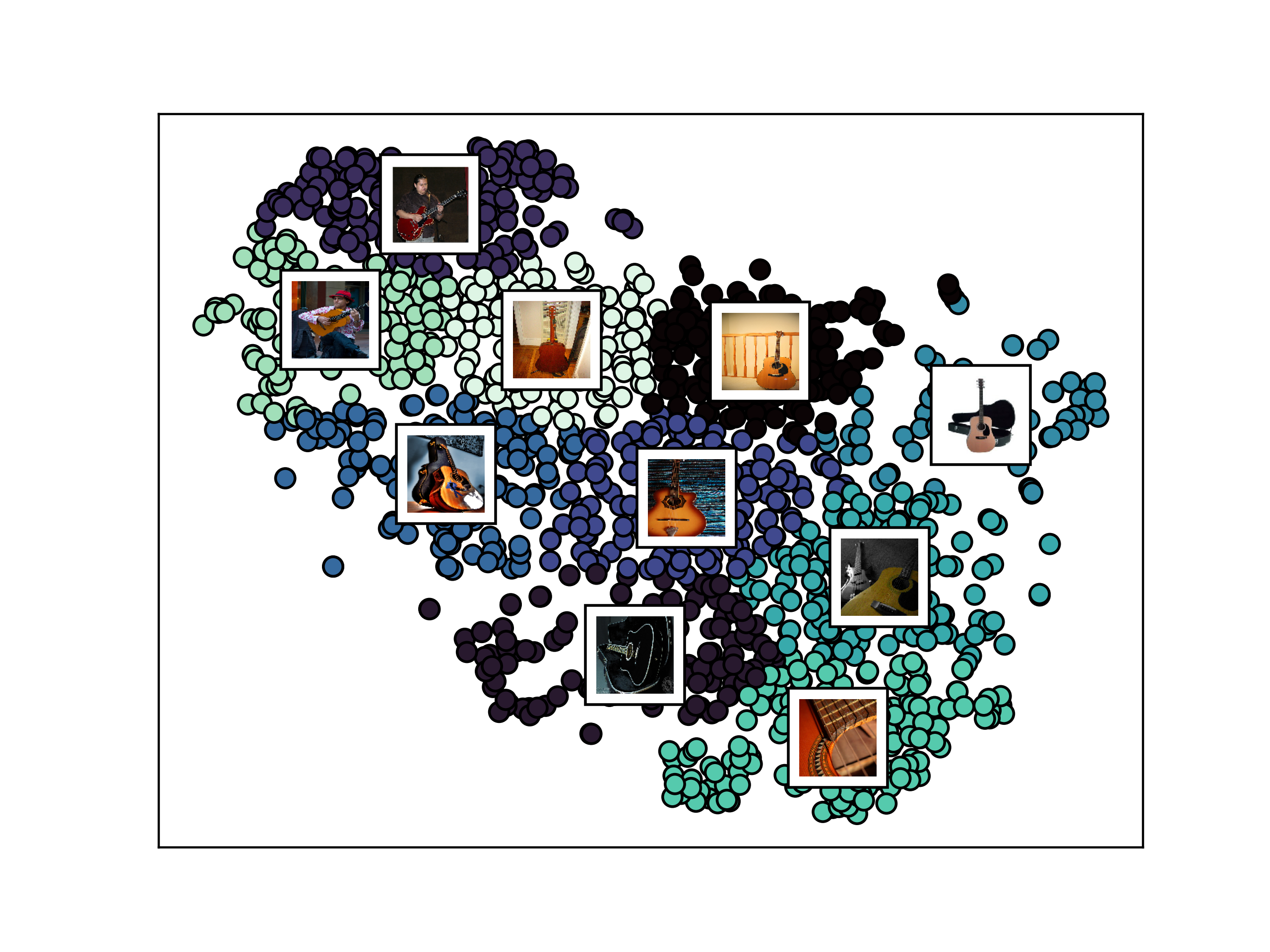}
        \caption*{b. Class ``acoustic guitar''}
        \label{fig:tsne_original_c}
    \end{minipage}
    \begin{minipage}[t]{.323\linewidth}
    \centering
        \includegraphics[trim=65px 40px 50px 45px, clip,width=\linewidth]{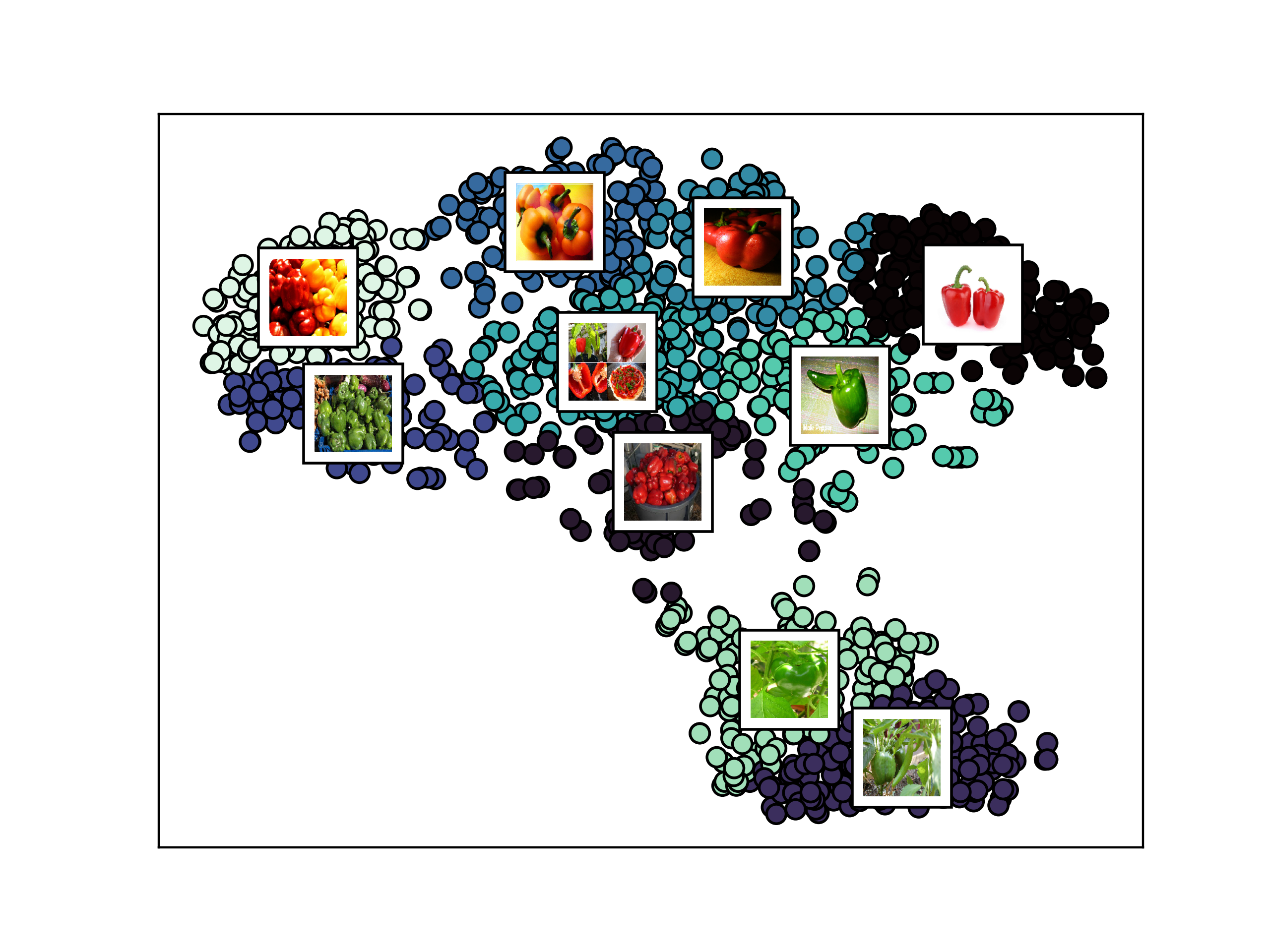}
        
        \includegraphics[trim=65px 40px 50px 45px, clip,width=\linewidth]{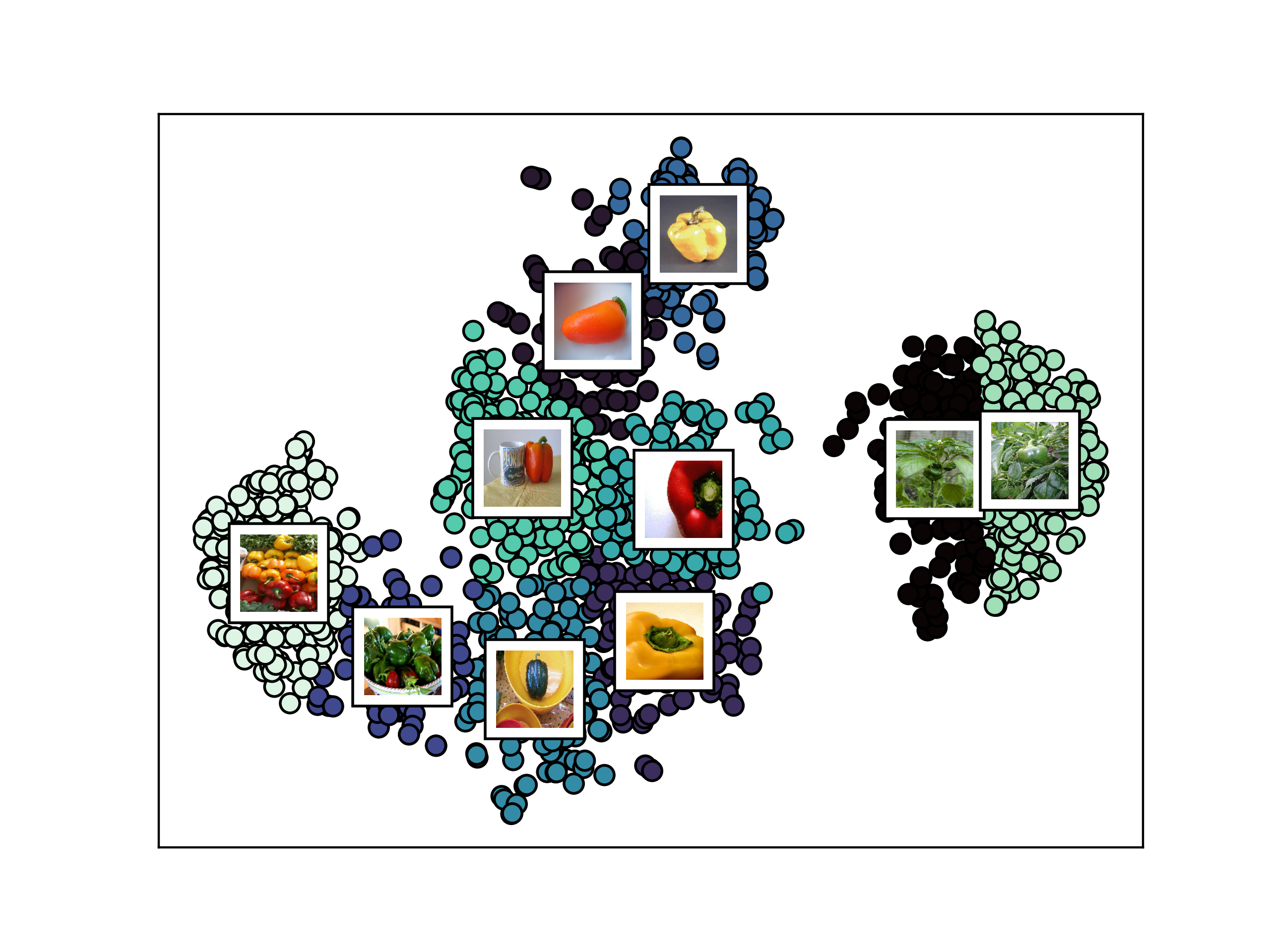}
        \caption*{c. Class ``bell pepper''}
        \label{fig:tsne_original_d}
    \end{minipage}
\setcounter{figure}{5}
\caption{\textbf{t-SNE feature embeddings for InceptionV3 with (bottom) and without (top) adaPool}. The ImageNet1K classes used are ``flamingo'', ``acoustic guitar'' and ``bell pepper''.}
\label{fig:tsne_kmeans}
\end{figure*}

\begin{figure*}[!htb]
\centering
    \includegraphics[width=\linewidth]{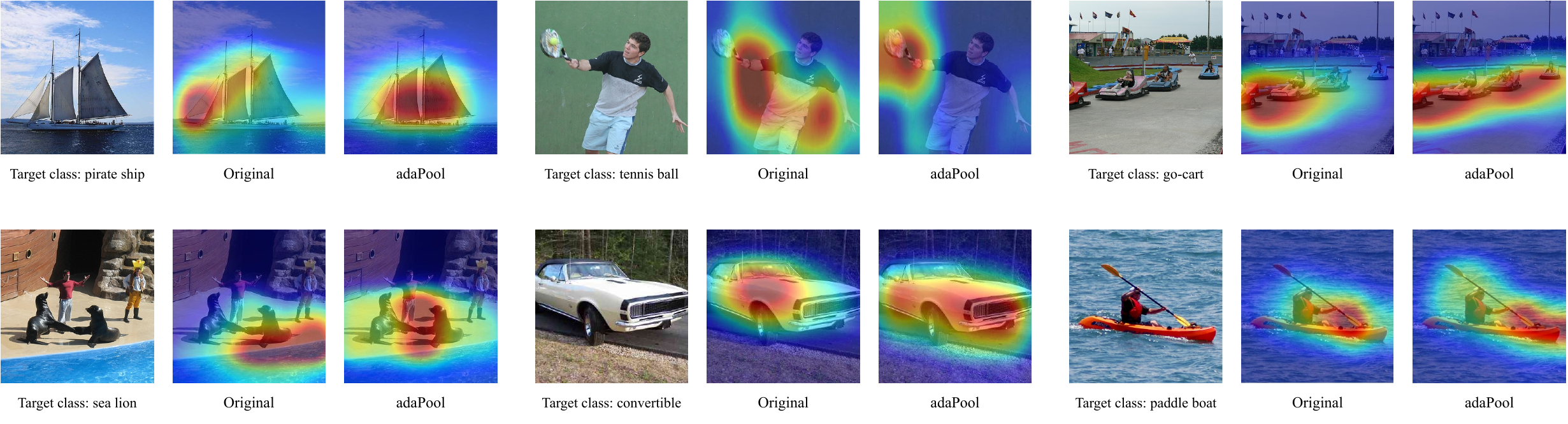}
\caption{\textbf{Saliency maps.} We compare maps of the visual saliency from two ResNet-50 models with the original max pooling and the proposed adaPool. Examples are sampled from the validation set of ImageNet1K. For each image we show the ground truth label.}
\label{fig:gradcam_ada}
\end{figure*}

\subsection{Qualitative visualizations}
\label{sec:ablations::visualizations}

To better understand the effect of adaPool in the feature extraction process, we compute saliency maps using Grad-CAM \cite{selvaraju2017grad} to visualize the salient regions for the original and adaPool-substituted networks, shown in Figure~\ref{fig:gradcam_ada}. We use a fixed ResNet-50 model from Table~\ref{tab:ImageNet_scratch} and sample examples from the ImageNet classes ``pirate ship'', ``tennis ball'', ``go-cart'', ``sea lion'', ``convertible'' and ``paddle boat''. 

For cases such as ``go-cart'' and ``sea lion'' where multiple objects of the class appear in the image, the adaPool-based network produces saliency maps that better fit their regions. Because details regarding the input are better preserved, the model focuses more on regions containing more descriptive features of the class, for example the sails in the ``pirate ship'' example or the racket and ball for ``tennis ball''.

Additionally, in Figure~\ref{fig:tsne_kmeans} we provide t-SNE \cite{van2008visualizing} visualizations for the feature embeddings of the original and adaPool-replaced InceptionV3. We follow the same recipe as in \cite{stergiou2021refining} and reduce the dimensionality to 50 channels with PCA. Overall, feature embeddings for similar examples are shown to be mapped somewhat closer on the adaPool-enabled network. For example, there is a clearer distinction between the color of the peppers for the class ``bell pepper'' as well as a distinction between multiple or single peppers in an image. 

\section{Conclusion}
\label{sec:conclusion}

In this paper, we have proposed adaPool, a pooling method for the preservation of informative features based on adaptive exponential weighting. It is a regionally-adaptive method that uses the parameterized fusion of the exponential maximum eMPool and exponential average eDSCWPool. The weights of adaPool can be used to invert the pooling operation (adaUnPool), to achieve upsampling.

We have tested our approach on image and video classification, image similarity, object detection, image and frame super-resolution tasks, as well as frame interpolation. The experiments consistently demonstrate the merits of our proposed approach when faced with various challenges such as capturing global and local information, or to consider 2D image data and 3D video data. Over all downstream tasks, and using a variety of network backbones and experiment settings, adaPool systematically outperforms any other method while computational latencies and memory use remain modest. Based on these extensive experiments, we believe adaPool is a good alternative for currently popular pooling operators.

\appendices

\section{}
\label{ap:idw}

In this appendix, we provide more details on Inverse Distance Weighting (IDW) pooling (Section~\ref{ap:A::IDW}), a motivation for our use of the Dice-S\o rensen Coefficient (DSC, Section~\ref{ap:B::DSC}), a comparison with other soft average methods (Section~\ref{sec:ablations::average_methods}), and a description of the computational complexity of our implementation (Section~\ref{ap:A:compute}).

\subsection{Inverse Distance Weighting pooling}
\label{ap:A::IDW}

\begin{figure}[h]
\includegraphics[width=\linewidth]{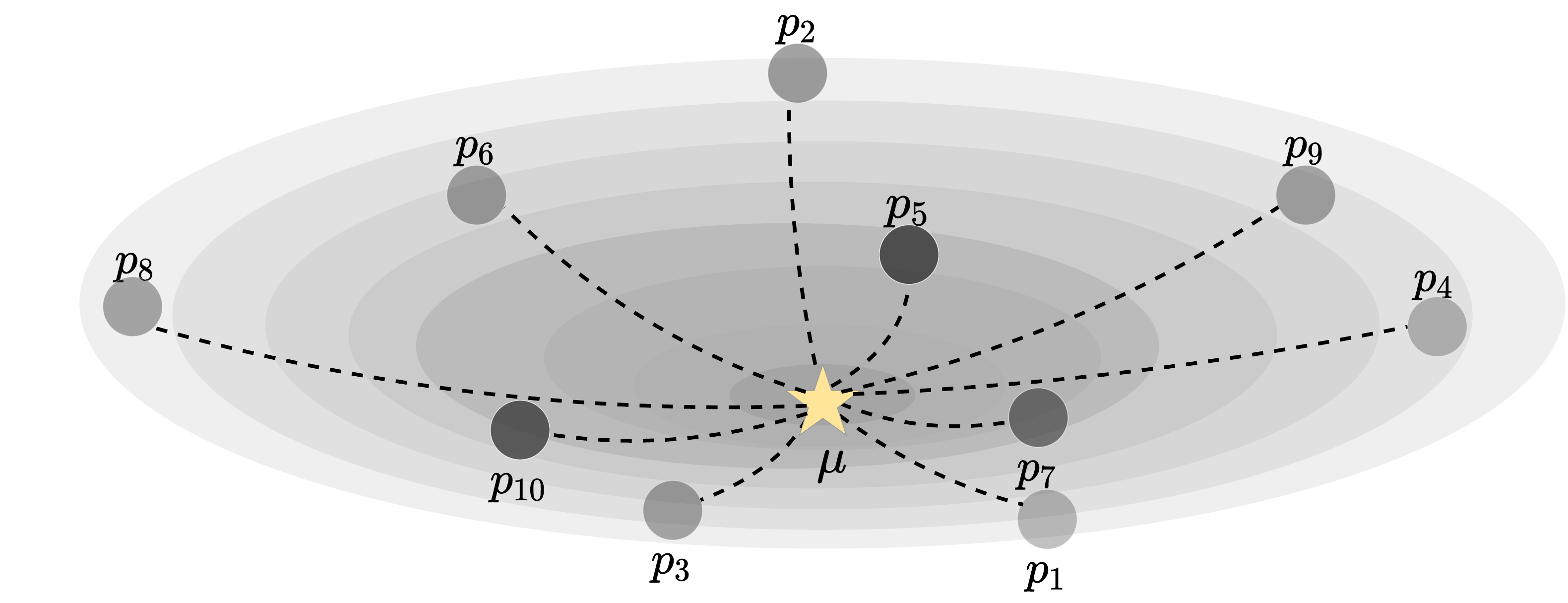}
\caption{\textbf{Inverse Distance Weighting}. Given multiple points $\{p_{1},...,p_{n}\}$ in a feature space and their mean ($\mu$), their weights are equal to the inverse of their distance divided by their sum.}
\label{fig:idw_avg}
\end{figure}

To assign a weight value, IDW relies on the measured observation distances within the region. A visual representation of this weighting process is shown in Figure~\ref{fig:idw_avg}.

To overcome the limitations of uniformly-weighted region averaging, we adapt IDW for pooling, which we term \textit{IDWPool}. Our results in Section~\ref{sec:experiments} use the Euclidean distance ($L_2$) between the mean and the individual activations. We also provide an overview alongside results for alternative distance functions in the following sections. In comparison to uniformly-weighted averaging, IDWPool produces normalized results with higher weights for feature activation vectors that are geometrically closer to the mean. This also applies to the calculation of the gradients, and reduces the effect of outliers, providing a better representative update rate based on feature activation relevance. In that aspect, IDWPool works differently than the common approach of averaging all activations in which the output activation is not regularized.

Although IDWPool can provide an improvement over uniformly-weighted averaging, we argue that weighted averaging based on distance is sub-optimal over multi-dimensional spaces. One of the main drawbacks of a naive IDWPool implementation is that the $L_1$ or $L_2$ distance between the feature activation vector and the average over the region are calculated based on the mean, sum or maximum per-channel pair. The resulting distance is unbounded since the pair-wise distances are also unbounded. In addition, the calculated distance is sensitive to channel pair outliers. The effect of this is visible with the pixel artifacts of the inverse distance weighting approaches in Figure~\ref{fig:avg_methods}. When using distance methods, the computed distance in certain channels can be significantly larger than in others. This creates the problem of weights that are nearing zero ($ \underset{IDW}{w(\mathbf{\overline a_c},\mathbf{a}_{j,c})} \rightarrow 0$).

\begin{table}[t]
\caption{\textbf{Distance functions for vectors}. All methods can be applied to multi-dimensional vector volumes.}
\label{tab:distance_eqs}
\centering
\begin{tabular}{|  m{1.6cm}  m{4cm} |}\hline
\vspace{-2ex}Manhattan ($L_{1}$) & {\begin{flalign}
\underset{L_{1}}{d} = \sum\limits_{c \in \mathbf{C}}||\mathbf{\overline a}_{c} - \mathbf{a}_{i,c}||&&
\label{eq:l_1}\end{flalign}}\\[-2ex]

\hline
\vspace{-2ex}Euclidean ($L_{2}$) & {\begin{flalign}
\underset{L_{2}}{d} = \sum\limits_{c \in \mathbf{C}}\sqrt{||\mathbf{\overline a}_{c} - \mathbf{a}_{i,c}||^{2}}&&
\label{eq:l_2}\end{flalign}}\\[-2ex]

\hline
\vspace{-2ex}Huber \cite{huber1992robust} & {\begin{flalign}
\underset{Hub}{d} = \begin{dcases}
\frac{d_{L_1} ^{2}}{2}, \, if \; d_{L_2} \leq \delta \\
\delta \centerdot (d_{L_{1}} - \frac{\delta}{2})
\end{dcases}&&
\label{eq:huber}\end{flalign}}\\[-2ex]

\hline
\vspace{-2ex}Chebyshev \cite{van2005classification} & {\begin{flalign}
\underset{L_{Che}}{d} = \underset{c \in C}{max}\;d_{L_{1}}&&
\label{eq:chebyshev}\end{flalign}}\\[-2ex]

\hline
\vspace{-2ex}Gower \cite{gower1971general} & {\begin{flalign}
\underset{L_{Gow}}{d} = \frac{1}{C} \centerdot d_{L_{1}} &&
\label{eq:gower}\end{flalign}}\\

\hline
\end{tabular}
\vspace{1em}
\end{table}

\subsection{Coefficient-based methods}
\label{ap:B::DSC}

\begin{figure}[t]
\includegraphics[width=\linewidth]{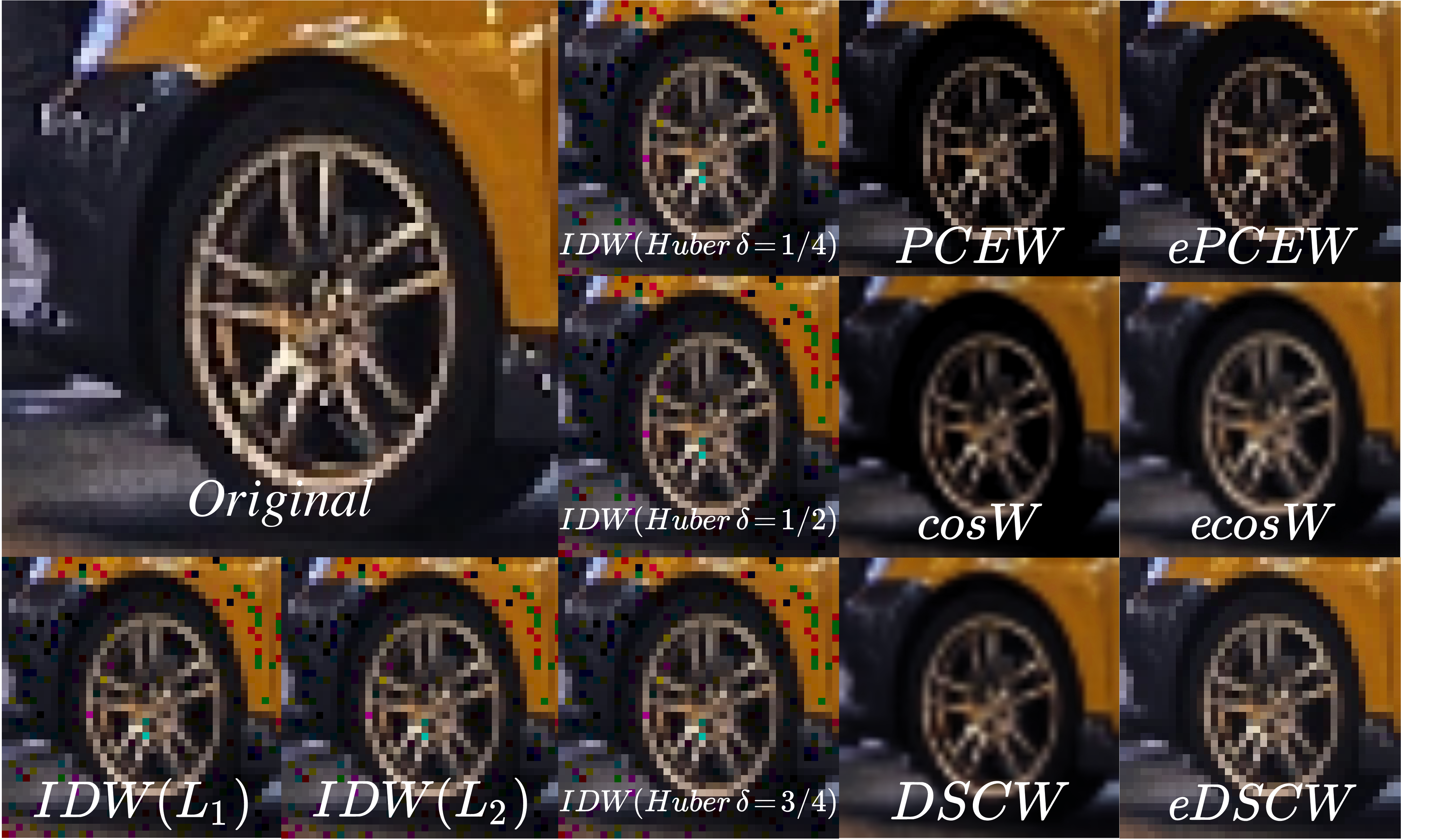}
\caption{\textbf{Instances of average distance/similarity weighting methods}. Distance kernel weights based on IDW \cite{shepard1968two} with various inverse distance functions. Similarity kernel weights based on (\textit{e})PCEW, (\textit{e})cosW and (\textit{e})DSCW.}
\label{fig:avg_methods}
\end{figure}

We have considered other similarity-based methods to find the relevance of two volumes of vectors \cite{cha2007comprehensive}. Apart from the cosine similarity, the Kumar and Hassebrook Peak-to-correlation energy (PCE) \cite{kumar1990performance} can be applied to vector volumes (as shown in Table~\ref{tab:similarity_eqs}). We present the differences in the pooling quality based on different similarity methods in Figure~\ref{fig:avg_methods}. Considering the aforementioned shortfalls of cosine similarity, our use of DSC over PCE is primarily due to PCE's non-monotonic nature and value distribution \cite{kumar1990performance}.

\begin{table}
\caption{\textbf{Similarity functions for vectors}. All methods can be directly applied to multi-dimensional vector volumes.}
\label{tab:similarity_eqs}
\centering
\begin{tabular}{|  m{2cm}  m{5cm} |}\hline
Cosine & {\begin{flalign}
\underset{cos}{S} = \frac{\sum\limits_{c \in \mathbf{C}} \mathbf{\overline a} \centerdot \mathbf{a}_{i,c}}{\sqrt{\sum\limits_{c \in C}\mathbf{\overline a}_{c}^{2}} \centerdot \sqrt{\sum\limits_{c \in C}\mathbf{a}_{c}^{2}}}&&
\label{eq:cosine}\end{flalign}}\\[-2ex]
\hline
PCE & {\begin{flalign}
\underset{PCE}{S} = \frac{\sum\limits_{c \in \mathbf{C}} \mathbf{\overline a} \centerdot \mathbf{a}_{i,c}}{\sum\limits_{c \in C}\mathbf{\overline a}_{c}^{2} \! + \! \! \sum\limits_{c \in C}\mathbf{a}_{c}^{2} \! - \! \! \sum\limits_{c \in C}\mathbf{\overline a}_{c} \centerdot \mathbf{a}_{c}}&&
\label{eq:lpce}\end{flalign}}\\
\hline
DSC & {\begin{flalign}
\underset{DSC}{S} = \sum\limits_{c \in \mathbf{C}}\frac{2 \centerdot ||\mathbf{\overline a}_{c} \centerdot \mathbf{a}_{i,c}||}{||\mathbf{\overline a}_{c}||^{2} + ||\mathbf{a}_{i,c}||^{2}}&&
\label{eq:dsc}\end{flalign}}\\
\hline
\end{tabular}
\end{table}

\subsection{Comparison with alternative soft average methods}
\label{sec:ablations::average_methods}

\begin{table}[ht]
\caption{\textbf{ImageNet1K classification with distance- and similarity-based pooling alternatives on ResNet-18}. Distance-based methods are denoted by IDW, while similarity-based methods are denoted with Sim. Best results in \textbf{bold}.}
\label{tab:avg_based_tests}
\begin{center}
    \centering
    \resizebox{.8\linewidth}{!}{%
    \begin{tabular}{l|ll|cc}
    \cline{1-5}
    \multicolumn{3}{c|}{Method} & top-1 & top-5\\[0.1em] 
    \hline
    & \multicolumn{2}{l|}{Original (Baseline)} & 69.76 & 89.08 \\[0.1em]
    \hline
    \multirow{7}{*}{\rotatebox{90}{IDW}} 
    & \multicolumn{2}{l|}{$L_{1}$} & 69.94 (\textcolor{applegreen}{+0.18}) & 89.24 (\textcolor{applegreen}{+0.16}) \\[0.1em]
    & \multicolumn{2}{l|}{$L_{2}$} & 70.02 (\textcolor{applegreen}{+0.23}) & 89.28 (\textcolor{applegreen}{+0.20}) \\[0.1em]
    & \multirow{3}{*}{\rotatebox{90}{\footnotesize Huber \cite{huber1992robust}}} & $\delta=1/4$ & 70.11 (\textcolor{applegreen}{+0.35}) & 89.33 (\textcolor{applegreen}{+0.25}) \\[0.1em]
    & & $\delta=1/2$ & 70.09 (\textcolor{applegreen}{+0.33}) & 89.27 (\textcolor{applegreen}{+0.19}) \\[0.1em]
    & & $\delta=3/4$ & 70.13 (\textcolor{applegreen}{+0.37}) & 89.32 (\textcolor{applegreen}{+0.24}) \\[0.1em]
    & \multicolumn{2}{l|}{Chedyshev} & 69.96 (\textcolor{applegreen}{+0.20}) & 89.20 (\textcolor{applegreen}{+0.12}) \\[0.1em]
    & \multicolumn{2}{l|}{Gower} & 69.58 (\textcolor{my_red}{-0.18}) & 88.94 (\textcolor{my_red}{-0.14}) \\[0.1em]
    \hline
    \multirow{4}{*}{\rotatebox{90}{Sim.}} 
    & \multicolumn{2}{l|}{Cosine} & 70.45 (\textcolor{applegreen}{+0.69}) & 89.44 (\textcolor{applegreen}{+0.36}) \\[0.2em]
    & \multicolumn{2}{l|}{PCE} & 70.54 (\textcolor{applegreen}{+0.78}) & 89.51 (\textcolor{applegreen}{+0.43}) \\[0.2em]
    & \multicolumn{2}{l|}{DSC} & \textbf{70.66} (\textcolor{applegreen}{+0.90}) & \textbf{89.77} (\textcolor{applegreen}{+0.69}) \\[0.2em]
    
    \end{tabular}
    }
\end{center}%
\end{table}

To evaluate the effect of different distance and similarity measures for average-approximating pooling in image classification performance, we use a ResNet-18 as backbone. We set as baseline the original ResNet-18 with maximum pooling.

The results in Table~\ref{tab:avg_based_tests} show negligible differences between distances in IDW pooling. Huber-based pooling shows small top-1 accuracy improvements, in the range of +(0.10--0.19)\% over $L_{1}$, $L_{2}$ and Chebyshev distance-weighting. A slight performance reduction is observed with the Gower method. This could be because of the production of small weight values as Gower uses the $L_{1}$ distance divided by the number of channels (Equation~\ref{eq:gower}).

Compared to distance approaches, similarity measures show a larger increase over the baseline model. This can be attributed to the sparsity of the per-input volumes. Considering the relatively small size of the kernel ($k \! \times \! k$) and the high-dimensional spaces they are represented in, distances between points and their mean are large \cite{domingos2012few}. The Dice-S\o rensen coefficient is most effective with 70.66\% and 89.77\% top-1 and top-5 accuracies. Increases are observed by the exponent of DSC in \textit{e}DSCWPool shown in Table~\ref{tab:ImageNet_scratch}, with 70.79\% top-1 and 90.16\% top-5 accuracies.

\subsection{Ablations over $\beta$ parameterization alternatives}
\label{sec:ablations::beta_param}

As adaPool introduces additional parameters. Therefore, we evaluate if the observed gains in performance are indeed due to improved information retainment or simply due to the inclusion of more parameters. We use three different $\beta$ sizes: a single $|\beta|=1$ parameter shared across each location, our proposed mask $|\beta|=H' \times W'$ for individual parameters across each location, and a channel-wise mask $|\beta|=H' \times W' \times C$ for both location and channel-based parameters. We present results on ResNet-50 and DenseNet-161 in Table~\ref{tab:beta_params}. We observe a difference between our proposed mask-based $\beta$ and the largely parameterized channel-wise $\beta$ on both models, with 1.01\% in ResNet-50 and 1.28\% in DenseNet-121. The results suggests that improvements in performance are not solely dependent on the inclusion of additional parameters. The channel-wise $\beta$ underperforms compared to the other non-channel-wise parameterization approaches. This suggests that the pooling approach is better suited for data with larger channel and feature dependencies. Our proposed approach introduces only a small fraction of additional parameters compared to the parameters used by most models, with +3.1K on ResNets and +4.2K on DenseNets (see Table~\ref{tab:flops_params}). We conclude that the observed performance improvements are strongly related to the design of adaPool instead of the additional parameters.

\begin{table}[t]
\caption{\textbf{AdaPool $\beta$ parameterization alternatives on ImageNet1K} for ResNet-50 and DenseNet-121. Best results and settings in \textbf{bold}.}
\label{tab:beta_params}
\centering
\resizebox{.75\linewidth}{!}{
\begin{tabular}{l|lll}\hline
\multicolumn{1}{c|}{Method} & top-1 & Params & FLOPs \tstrut \bstrut \\ \hline
\multicolumn{4}{l}{ResNet-50}\tstrut \\
Baseline (AvgPool) & 76.15 & 25.6M & 4.14G \\
$\beta$ single  & 77.76 & +1 & +0.8M \\
$\beta$ mask (proposed) & \textbf{78.42} & +3.1K & +0.8M \\
$\beta$ channel-wise & 77.41 & +198.5K & +0.8M \bstrut \\
\hline
\multicolumn{4}{l}{DenseNet-121} \tstrut \\
Baseline (AvgPool) & 74.65 & 8.6M & 2.9G \\
$\beta$ single  & 76.41 & +4 & +1.5M \\
$\beta$ mask (proposed) & \textbf{77.29} & +4.2K & +1.5M \\
$\beta$ channel-wise & 76.13 & +0.5M & +1.5M \\
\end{tabular}
}
\end{table}

\begin{table}[t]
\caption{\textbf{Parameters and FLOPs} overhead with the inclusion of adaPool per family of architectures.}
\label{tab:flops_params}
\centering
\resizebox{.7\linewidth}{!}{
\begin{tabular}{l|cc}\hline
\multicolumn{1}{c|}{Arch.} & Params (K) & FLOPs (M) \\ \hline
ResNets & +3.1 & +0.8 \\
InceptionV3 & +3.5 & +1.3 \\
DenseNets & +4.2 & +1.5 \\
\hline
\end{tabular}
}
\end{table}

\subsection{Computational description}
\label{ap:A:compute}

Our implementation is in CUDA and thus allows the native run on GPUs, providing inference times close to those of native methods such as average and maximum pooling. Due to the parallelization capabilities of both exponential maximum and average pooling methods, running times are close to those of average pooling with $\mathcal{O}(2)$ and $\mathcal{O}(3)$ respectively, as operations can be performed in parallel over the kernel region matrix. In contrast, max pooling has $\mathcal{O}(n)$ computational complexity, due to the sequential consideration of each input within the region in order to discover the maximum.

Both eMPool and eDSCWPool are on par with average and maximum pooling due to CUDA's memory reduction through data partitioning with tiling. In addition, both can be implemented through fused multiply-adds (FMA) that significantly improve performance on CUDA-enabled devices \cite{lee2010debunking}.

\section*{Acknowledgment}
The authors thank the Netherlands Organization for Scientific Research (NWO) for the support through TOP-C2 grant ``Automatic recognition of bodily interactions'' (ARBITER).

\ifCLASSOPTIONcaptionsoff
  \newpage
\fi



\bibliographystyle{IEEEtran}
\bibliography{IEEEfull}
%

%

\begin{IEEEbiography}[{\includegraphics[width=1in,height=1.25in,clip,keepaspectratio]{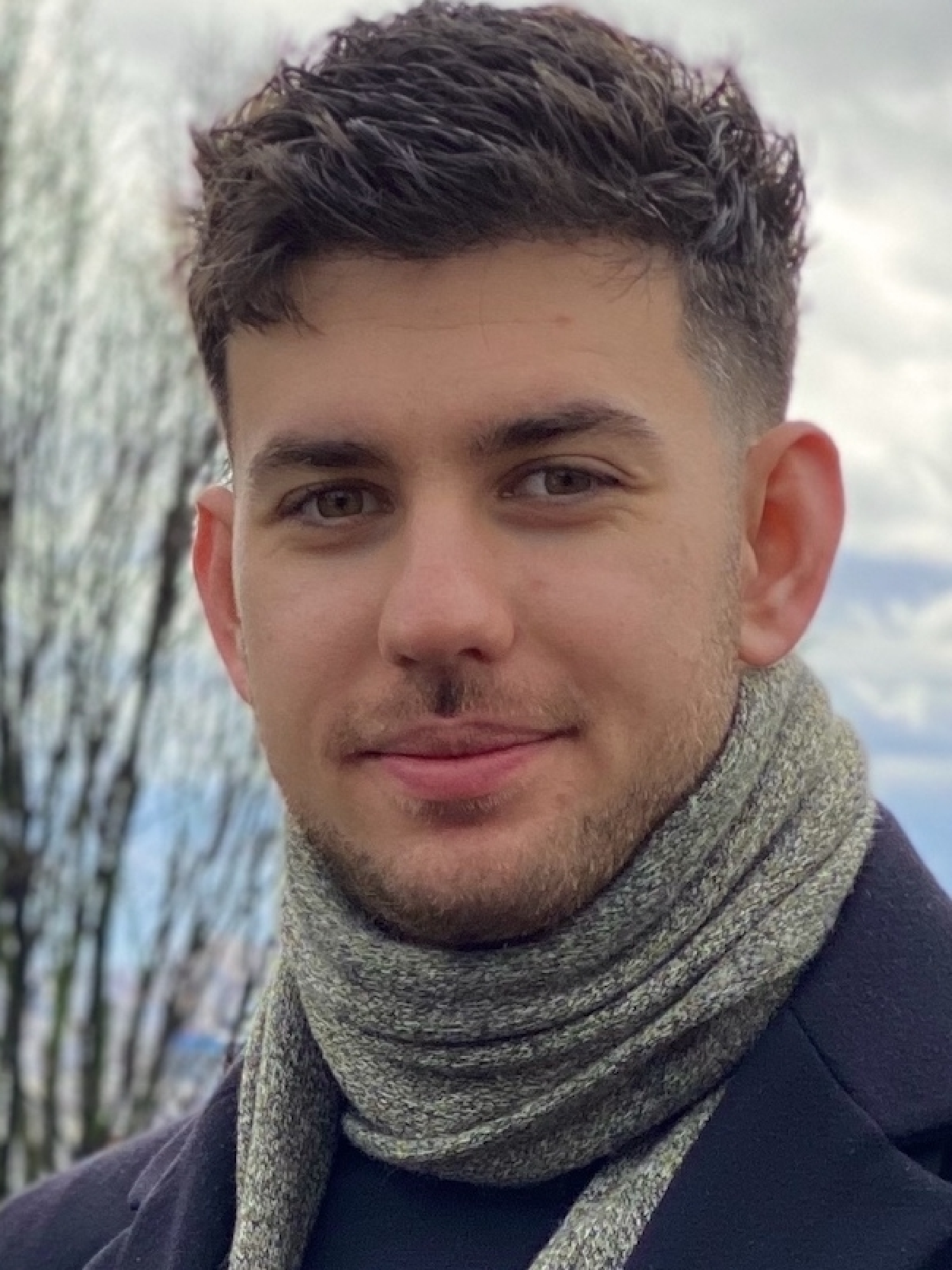}}]%
{Alexandros Stergiou}
(Student Member, \textit{IEEE}) received his Ph.D. degree in Computer Science from Utrecht University's Department of Information and Computing Sciences (2021). He obtained his B.Sc. and M.Sc degrees in Computer Science from the University of Essex. He is currently a Research Associate at University of Bristol's Department of Computer Science. His research interests include recognition and prediction of human actions from videos and deep learning model explainability. 
\end{IEEEbiography}

\begin{IEEEbiography}[{\includegraphics[width=1in,height=1.25in,clip,keepaspectratio]{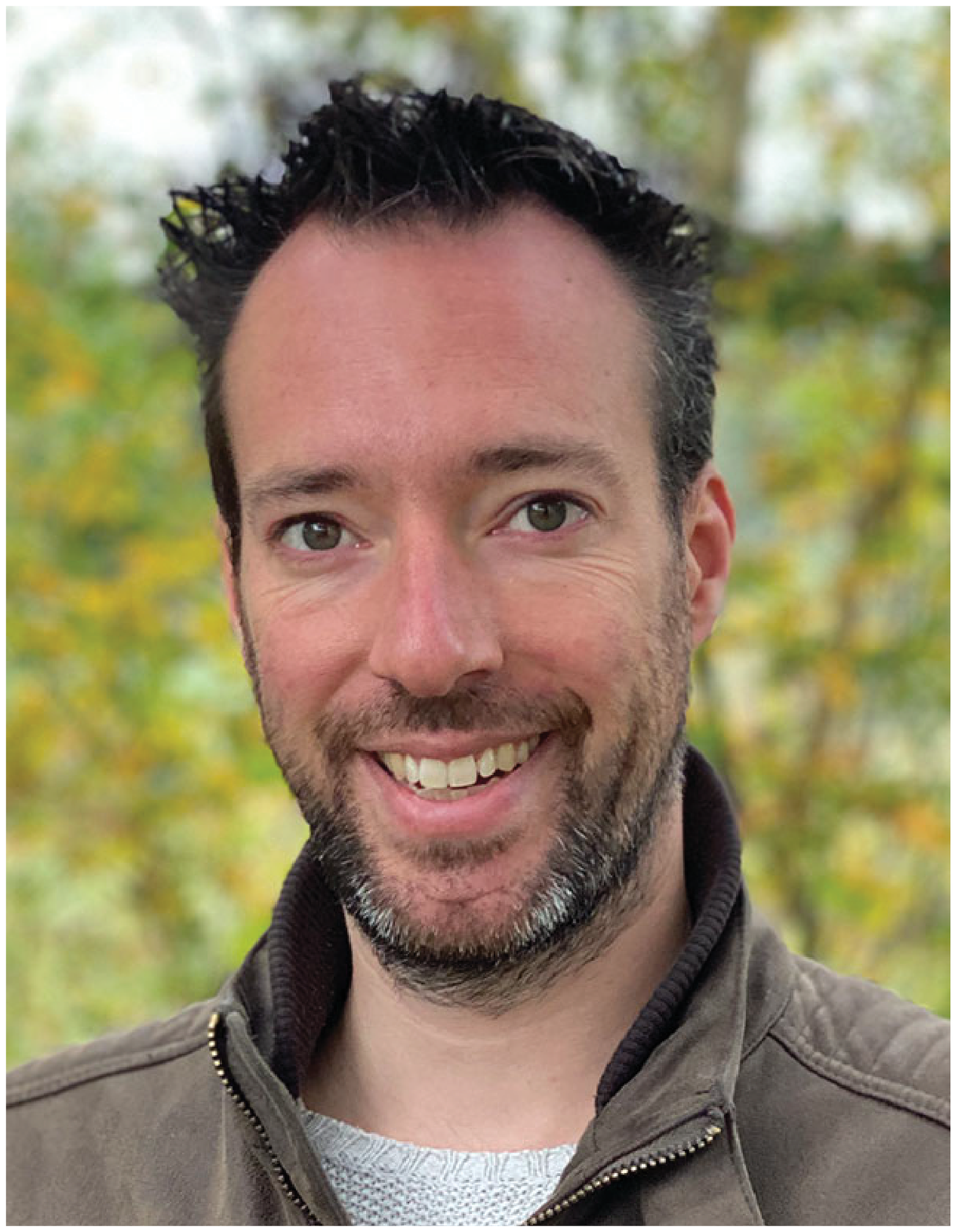}}]%
{Ronald Poppe} (Senior Member, \textit{IEEE})
received his Ph.D. in Computer Science from the University of Twente, the Netherlands (2009). He was a visiting researcher at the Delft University of Technology, Stanford University and University of Lancaster. He is currently an associate professor at the Department of Information and Computing Sciences of Utrecht University. His research interests include modeling of visual attention and the analysis of human (interactive) behavior from video.
\end{IEEEbiography}




\end{document}